\documentclass[journal]{IEEEtran}
\pdfoutput=1
\usepackage{amsmath,amsfonts}
\usepackage{algorithmic}
\usepackage{algorithm}
\usepackage{array}
\usepackage[caption=false,font=normalsize,labelfont=sf,textfont=sf]{subfig}
\usepackage{textcomp}
\usepackage{stfloats}
\usepackage{url}
\usepackage{verbatim}
\usepackage{graphicx}
\usepackage{cite}
\usepackage{amssymb}
\usepackage{xcolor}
\usepackage{booktabs}
\usepackage{multirow}
\usepackage{pifont}
\usepackage{makecell}
\usepackage{url}

\newtheorem{theorem}{\bf Theorem}
\newtheorem{definition}{\bf Definition}
\newtheorem{lemma}{Lemma}

\begin{document}

\title{RNN-Guard: Certified Robustness Against Multi-frame Attacks for Recurrent Neural Networks}

\author{Yunruo~Zhang,~\IEEEmembership{}
        Tianyu~Du,~\IEEEmembership{}
        Shouling~Ji,~\IEEEmembership{Member,~IEEE,}
        Peng~Tang,~\IEEEmembership{}
        and~Shanqing~Guo
\thanks{Manuscript received XX, YEAR; revised YY, YEAR. (Corresponding authors: Shouling Ji; Shanqing Guo.)}%
\thanks{Yunruo Zhang, Peng Tang, and Shanqing Guo are with the School of Cyber Science and Technology at Shandong University, Qingdao, Shandong 266237, China (e-mail: zhangyunruo@mail.sdu.edu.cn; tangpeng@sdu.edu.cn; guoshanqing@sdu.edu.cn).}%
\thanks{Tianyu Du is with the College of the Information and Science Technology, Penn State University, State College, PA 16801, U.S. (e-mail: tjd6042@psu.edu).}%
\thanks{Shouling Ji is with the College of Computer Science and Technology, Zhejiang University, Hangzhou, Zhejiang 310027, China, and also with the Binjiang Institute of Zhejiang University, Hangzhou, Zhejiang 310027, China. (e-mail: sji@zju.edu.cn).}%
}

\markboth{ }%
{ }


\maketitle

\begin{abstract}
It is well-known that recurrent neural networks (RNNs), although widely used, are vulnerable to adversarial attacks including one-frame attacks and multi-frame attacks. Though a few certified defenses exist to provide guaranteed robustness against one-frame attacks,  we prove that defending against multi-frame attacks remains a challenging problem due to their enormous perturbation space. In this paper, we propose the first certified defense against multi-frame attacks for RNNs called RNN-Guard. To address the above challenge, we adopt the perturb-all-frame strategy to construct perturbation spaces consistent with those in multi-frame attacks. However, the perturb-all-frame strategy causes a precision issue in linear relaxations. To address this issue, we introduce a novel abstract domain called InterZono and design tighter relaxations. We prove that InterZono is more precise than Zonotope yet carries the same time complexity. Experimental evaluations across various datasets and model structures show that the certified robust accuracy calculated by RNN-Guard with InterZono is up to 2.18 times higher than that with Zonotope. In addition, we extend RNN-Guard as the first certified training method against multi-frame attacks to directly enhance RNNs' robustness. The results show that the certified robust accuracy of models trained with RNN-Guard against multi-frame attacks is 15.47 to 67.65 percentage points higher than those with other training methods. 
\end{abstract}


\section{Introduction}\label{sec:introduction}
\IEEEPARstart{R}{ecurrent} neural networks have been widely applied in various applications, including natural language processing \cite{DBLP:conf/emnlp/LeiZWDA18}, automatic speech recognition \cite{DBLP:conf/icassp/GravesMH13}, and video processing \cite{DBLP:journals/tip/ZhaoLL19}.
However, due to the lack of robustness, RNNs are vulnerable to adversarial attacks \cite{DBLP:journals/corr/SzegedyZSBEGF13,DBLP:journals/corr/GoodfellowSS14,DBLP:conf/sp/Carlini017,DBLP:conf/iclr/MadryMSTV18}, where adversaries add small perturbations to clean inputs to induce the target model to misclassify the perturbed inputs (termed as adversarial examples).
The existence of adversarial attacks raises concerns about the safety of RNN-based applications.
To improve the robustness of RNNs, earlier researchers proposed many empirical defenses based on heuristics (e.g., adversarial training \cite{DBLP:journals/corr/GoodfellowSS14,DBLP:conf/cvpr/ShrivastavaPTSW17}).
However, empirical defenses can often be defeated by stronger attacks due to their lack of theoretical guarantees.
To provide guaranteed robustness, recently, a few \textit{robustness certification} methods (a.k.a. \textit{certified defenses}) \cite{DBLP:conf/icml/KoLWDWL19,DBLP:conf/ccs/DuJSZLSFYB021} have been proposed for RNNs to formally verify whether a given neighbor around the clean input contains any adversarial example.
Robustness certification methods can be further extended as certified training methods, which utilize the certification results for improving the model's robustness.

\begin{figure}[t]
    \centering
    \subfloat[]{
    \includegraphics[width=1.0in]{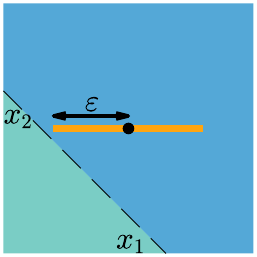}
    }
    \hfil
    \subfloat[]{
    \includegraphics[width=1.0in]{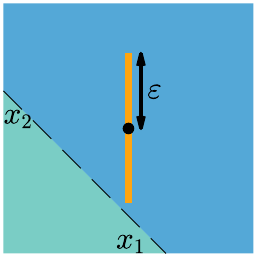}
    }
    \hfil
    \subfloat[]{
    \includegraphics[width=1.0in]{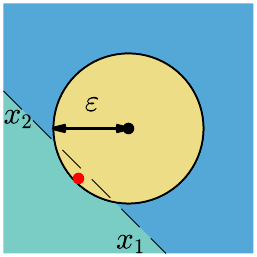}
    }
\caption{Examples of the perturbation space in one-frame attacks (the horizontal and vertical orange line) and multi-frame attacks (the yellow circle), where the black dot is a clean input with two frames (i.e., $x_1$ and $x_2$) and the red dot is an adversarial example in the multi-frame attack. (a) One-frame attack on $x_1$. (b) One-frame attack on $x_2$. (c) Multi-frame attack on $x_1$ and $x_2$.}
\label{fig:me}
\end{figure}

\begin{figure*}[t]
    \centering
    \includegraphics[width=\textwidth]{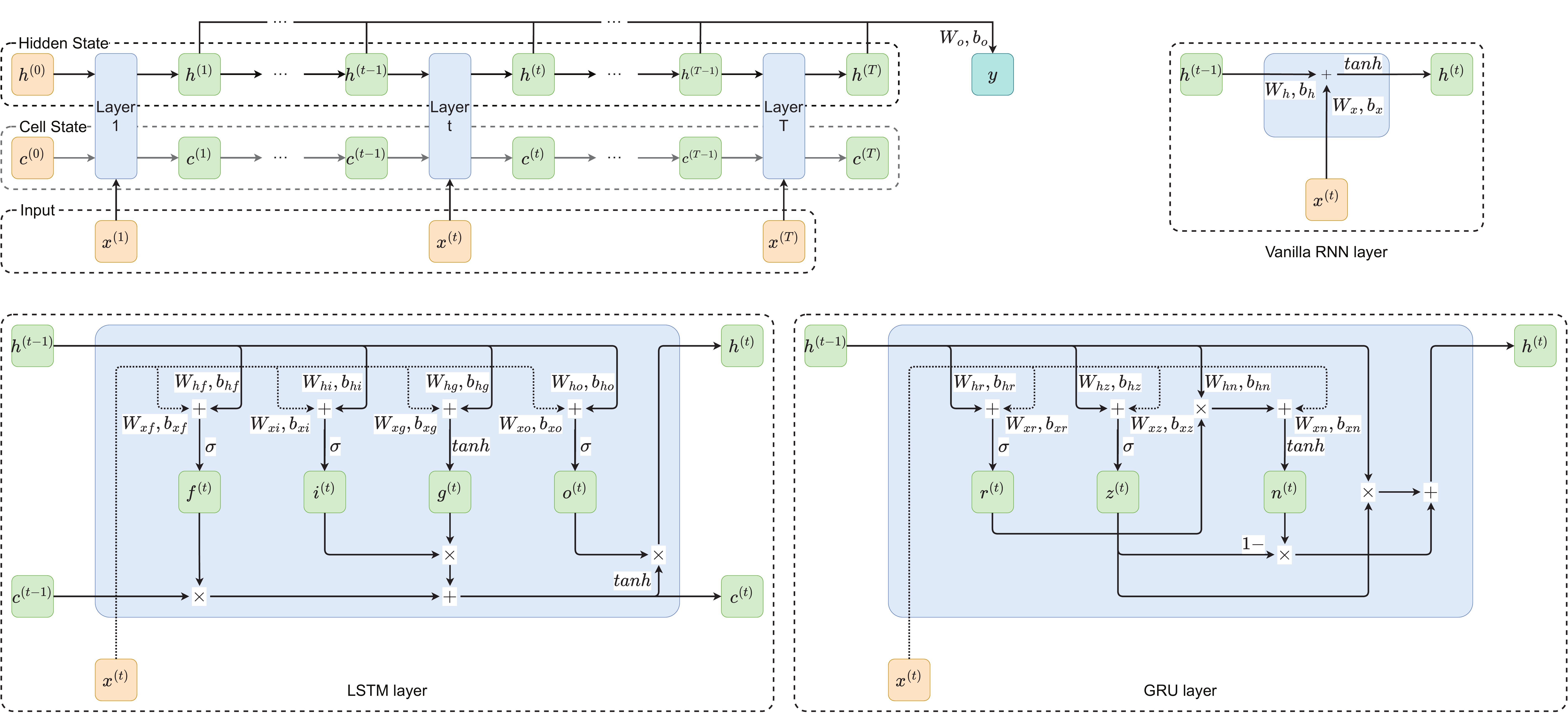}
    \caption{The recurrent neural network architecture for sequence classification. The cell state is unique to LSTM models.}
    \label{fig:rnn}
\end{figure*}

\textbf{Motivation.} 
Adversarial attacks on RNNs can be categorized into either \textit{one-frame attacks} where adversaries can only perturb one frame or \textit{multi-frame attacks} where adversaries can perturb multiple frames simultaneously. Previous certified defenses for RNNs focus on one-frame attacks. However, we discover a severe vulnerability in those works, i.e., they are vulnerable to multi-frame attacks. As shown in Fig. \ref{fig:me}, even if the model is certified to be robust around a clean sample against one-frame attacks, multi-frame adversaries can still find its adversarial example. Thus, defending against multi-frame attacks becomes an important and urgent problem.

\textbf{This work: Certified Defense against Multi-frame Attacks.} 
In this paper, we address this problem and present the first certified defense against multi-frame attacks called RNN-Guard. To capture all potential adversarial examples in multi-frame attacks, we adopt the perturb-all-frame strategy (i.e., to simultaneously perturb all frames), which results in a much larger perturbation space. However, the perturb-all-frame strategy causes a precision issue. Most certified defenses apply linear relaxations to handle non-linear functions. Generally speaking, relaxations in smaller spaces are more precise because a shorter curve is closer to a straight line than a longer one. Thus, the larger perturbation space results in less precise relaxations. The precision issue can cause imprecise results of robustness certification, i.e., too many robust samples are incorrectly proven to be non-robust. To address this issue, we introduce a new abstract domain called InterZono and design tighter relaxations for non-linear functions in RNN models. InterZono consists of a main domain and a support domain, which can reduce errors in relaxations by computing the intersection of the two domains.

We comprehensively evaluate the performance of RNN-Guard across various datasets and model structures.
Experimental results show that certified robust accuracy calculated by RNN-Guard with InterZono is up to 2.18 times higher than by RNN-Guard with Zonotope (a classic abstract domain used in many certified defenses), which indicates that InterZono is more precise, while they share the same efficiency.
To further demonstrate RNN-Guard's practicality, we apply RNN-Guard to certify the effectiveness of existing defenses.
The results show that those defenses lack robustness against multi-frame attacks, which reaffirms the exigency of this work.
Furthermore, we extend RNN-Guard as a certified training method to directly improve RNNs' robustness against multi-frame attacks.
The results show that the certified robust accuracy of models trained with RNN-Guard against multi-frame attacks is 15.47 to 67.65 percentage points higher than other training methods, which indicates that RNN-Guard is the best choice for defending against such attacks.
We also conduct adaptive evaluations, in which the results reconfirm that RNN-Guard is more effective against multi-frame attacks.

\textbf{Our Contributions.} Our main contributions are:
\begin{itemize}
    \item We discover a severe vulnerability in existing certified defenses for RNNs, i.e., they are challenged by multi-frame attacks due to their limited perturbation space. 
    We theoretically prove that defending against multi-frame attacks is more difficult than against one-frame attacks.
    \item To address this challenge, we propose a novel certified defense with the perturb-all-frame strategy for capturing all potential adversarial examples in multi-frame attacks.
    To the best of our knowledge, this is the first certified defense that focuses on multi-frame attacks.
    \item To address the precision issue caused by the perturb-all-frame strategy, we introduce an abstract domain called InterZono and design tighter relaxations. We theoretically prove that InterZono is more precise than Zonotope while carrying the same time complexity.
    \item Through extensive evaluations, we show that InterZono is more precise than Zonotope while sharing the same efficiency. Moreover, we demonstrate RNN-Guard’s superiority in improving RNN models' robustness against multi-frame attacks.
\end{itemize}

\section{Background}

\begin{figure*}[pt]
    \centering
    \subfloat[]{
    \includegraphics[width=2.2in]{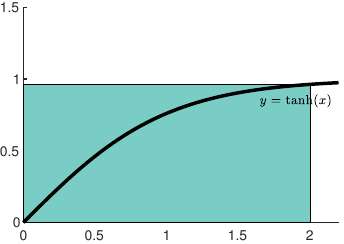}
    }
    \hfil
    \subfloat[]{ \label{fig:tanhb}
    \includegraphics[width=2.2in]{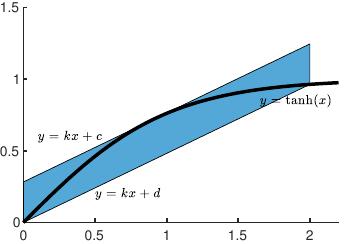}
    }
    \hfil
    \subfloat[]{
    \includegraphics[width=2.2in]{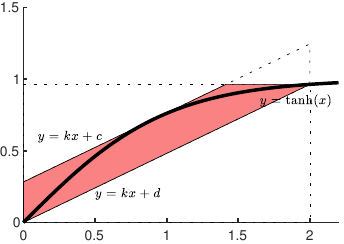}
    }
\caption{Comparison of relaxations for tanh. Our relaxed area is smaller than others. (a) Box-style relaxation. (b) Parallelogram-style relaxation. (c) Ours.}
\label{fig:tanh}
\end{figure*}

\subsection{Recurrent Neural Networks}
In this work, we apply RNNs to sequence classification tasks such as sentiment prediction and toxic content detection.
The RNN model receives a sequence of variable lengths (denoted by $T$) composed of frames such as words or tokens and performs classification into $C$ distinct classes.

Let $\boldsymbol{X} = [ \boldsymbol{x}^{(1)}, \boldsymbol{x}^{(2)}, \dots, \boldsymbol{x}^{(t)}, \dots, \boldsymbol{x}^{(T)} ]$ be the input sequence.
As shown in Fig. \ref{fig:rnn}, the RNN model updates a hidden state $\boldsymbol{h}^{(t)}$ at each time step according to the current frame $\boldsymbol{x}^{(t)}$ and the last hidden state $\boldsymbol{h}^{(t-1)}$:
$$
    \boldsymbol{h}^{(t)} = \text{layer} (\boldsymbol{x}^{(t)}, \boldsymbol{h}^{(t-1)}),
$$
where $t = 1, 2, \dots, T$. The initial hidden state $\boldsymbol{h}^{(0)}$ is usually set to zero.
In particular, LSTM models update an additional hidden state $\boldsymbol{c}^{(t)}$ called cell state.
The output vector $\boldsymbol{y} \in \mathbb{R}^C $ is calculated according to each hidden state:
$$
	\boldsymbol{y} = \boldsymbol{W}_{o} \cdot \boldsymbol{h}^{(1)} + \boldsymbol{W}_{o} \cdot \boldsymbol{h}^{(2)} + \dots + \boldsymbol{W}_{o} \cdot \boldsymbol{h}^{(T)} + \boldsymbol{b}_{o}.
$$
For simplicity, we replace it with $\boldsymbol{y} = \boldsymbol{W}_{o} \cdot \boldsymbol{h}^{(T)} + \boldsymbol{b}_{o}$.

We mainly consider three kinds of RNN models in this paper: vanilla RNN models, long short-term memory (LSTM) \cite{DBLP:journals/neco/HochreiterS97} models, and gated recurrent unit (GRU) \cite{DBLP:conf/ssst/ChoMBB14} models.
Each kind of model updates its hidden states differently.
Vanilla RNN models update $\boldsymbol{h}^{(t)}$ according to 
$$
    \boldsymbol{h}^{(t)} = \text{tanh} (\boldsymbol{W}_{x} \boldsymbol{x}^{(t)} + \boldsymbol{b}_{x} + \boldsymbol{W}_{h} \boldsymbol{h}^{(t-1)} + \boldsymbol{b}_{h}).
$$
LSTM models update $\boldsymbol{h}^{(t)}$ and $\boldsymbol{c}^{(t)}$ according to
\begin{align*}
	\boldsymbol{i}^{(t)} &= \sigma (\boldsymbol{W}_{xi} \boldsymbol{x}^{(t)} + \boldsymbol{b}_{xi} + \boldsymbol{W}_{hi} \boldsymbol{h}^{(t-1)} + \boldsymbol{b}_{hi}),\\
    \boldsymbol{f}^{(t)} &= \sigma (\boldsymbol{W}_{xf} \boldsymbol{x}^{(t)} + \boldsymbol{b}_{xf} + \boldsymbol{W}_{hf} \boldsymbol{h}^{(t-1)} + \boldsymbol{b}_{hf}),\\
    \boldsymbol{g}^{(t)} &= \text{tanh} (\boldsymbol{W}_{xg} \boldsymbol{x}^{(t)} + \boldsymbol{b}_{xg} + \boldsymbol{W}_{hg} \boldsymbol{h}^{(t-1)} + \boldsymbol{b}_{hg}),\\
    \boldsymbol{o}^{(t)} &= \sigma (\boldsymbol{W}_{xo} \boldsymbol{x}^{(t)} + \boldsymbol{b}_{xo} + \boldsymbol{W}_{ho} \boldsymbol{h}^{(t-1)} + \boldsymbol{b}_{ho}),\\
    \boldsymbol{c}^{(t)} &= \boldsymbol{f}^{(t)} \odot \boldsymbol{c}^{(t-1)} + \boldsymbol{i}^{(t)} \odot \boldsymbol{g}^{(t)},\\
    \boldsymbol{h}^{(t)} &= \boldsymbol{o}^{(t)} \odot \text{tanh} (\boldsymbol{c}^{(t)}),
\end{align*}
where $\odot$ is the Hadamard product.

GRU models update $\boldsymbol{h}^{(t)}$ according to
\begin{align*}
	\boldsymbol{r}^{(t)} &= \sigma (\boldsymbol{W}_{xr} \boldsymbol{x}^{(t)} + \boldsymbol{b}_{xr} + \boldsymbol{W}_{hr} \boldsymbol{h}^{(t-1)} + \boldsymbol{b}_{hr}),\\
    \boldsymbol{z}^{(t)} &= \sigma (\boldsymbol{W}_{xz} \boldsymbol{x}^{(t)} + \boldsymbol{b}_{xz} + \boldsymbol{W}_{hz} \boldsymbol{h}^{(t-1)} + \boldsymbol{b}_{hz}),\\
    \boldsymbol{n}^{(t)} &= \text{tanh} (\boldsymbol{W}_{xn} \boldsymbol{x}^{(t)} + \boldsymbol{b}_{xn} \\
    & \phantom{=} \ + \boldsymbol{r}^{(t)} \odot (\boldsymbol{W}_{hn} \boldsymbol{h}^{(t-1)} + \boldsymbol{b}_{hn})),\\
    \boldsymbol{h}^{(t)} &= (1 - \boldsymbol{z}^{(t)}) \odot \boldsymbol{n}^{(t)} + \boldsymbol{z}^{(t)} \odot \boldsymbol{h}^{(t-1)}.
\end{align*}

\subsection{Adversarial Attacks on RNNs} \label{sec:atk}
According to the number of frames in the input sequence that adversaries can perturb, adversarial attacks on RNNs are categorized into one-frame attacks or multi-frame attacks \cite{DBLP:conf/ccs/DuJSZLSFYB021}.

\subsubsection{One-frame Attacks}
One-frame attacks are adversarial attacks where the adversary can only perturb one frame and keep the others unchanged, which are considered in the previous works on certified defenses for RNNs \cite{DBLP:conf/icml/KoLWDWL19,DBLP:conf/ccs/DuJSZLSFYB021}.
For a clean input sequence $\boldsymbol{X}_c = [ \boldsymbol{x}_c^{(1)}, \boldsymbol{x}_c^{(2)}, \dots, \boldsymbol{x}_c^{(t)}, \dots, \boldsymbol{x}_c^{(T)} ]$, the one-frame adversarial example $\boldsymbol{X}_{adv}^{(t)}$ can be any sequence that satisfies:
\begin{gather*}
	\boldsymbol{X}_{adv}^{(t)} = [ \boldsymbol{x}_c^{(1)}, \boldsymbol{x}_c^{(2)}, \dots, \boldsymbol{x}_c^{(t-1)}, \boldsymbol{x}_{adv}^{(t)}, \boldsymbol{x}_c^{(t+1)}, \dots, \boldsymbol{x}_c^{(T)} ], \\
	\Vert \boldsymbol{x}_c^{(t)} - \boldsymbol{x}_{adv}^{(t)} \Vert_p \le \varepsilon, \boldsymbol{f}_{p}(\boldsymbol{X}_{adv}^{(t)}) \ne \boldsymbol{f}_{p}(\boldsymbol{X}), t = 1, 2, \dots, T,
\end{gather*}
where $\boldsymbol{f}_{p}$ returns the target model's prediction on the input. 

However, perturbing one frame is not always enough to achieve a successful attack.
Thus, to ensure a high success rate, most attacks in practice (e.g., DeepWordBug \cite{DBLP:conf/sp/GaoLSQ18} and BERT-Attack \cite{DBLP:conf/emnlp/LiMGXQ20}) usually try to perturb more frames together until an adversarial example is found, which are referred to as multi-frame attacks.

\subsubsection{Multi-frame Attacks}
Multi-frame attacks are adversarial attacks where the adversary can perturb multiple (partial or all) frames in the input sequence.
The perturbation on each perturbed frame is also constrained within a given length.
For example, the worst-case adversarial example $\boldsymbol{X}_{adv}$ where all frames are perturbed satisfies:
\begin{gather*}
	\boldsymbol{X}_{adv} = [ \boldsymbol{x}_{adv}^{(1)}, \boldsymbol{x}_{adv}^{(2)}, \dots, \boldsymbol{x}_{adv}^{(t-1)}, \boldsymbol{x}_{adv}^{(t)}, \boldsymbol{x}_{adv}^{(t+1)}, \dots, \boldsymbol{x}_{adv}^{(T)} ], \\
	\Vert \boldsymbol{x}_c^{(t)} - \boldsymbol{x}_{adv}^{(t)} \Vert_p \le \varepsilon, \boldsymbol{f}_{p}(\boldsymbol{X}_{adv}^{(t)}) \ne \boldsymbol{f}_{p}(\boldsymbol{X}), t = 1, 2, \dots, T.
\end{gather*}

\subsection{Zonotope Certification}
This work builds upon the Zonotope \cite{DBLP:conf/cav/GhorbalGP09} abstract domain, which was used in many certified defenses \cite{DBLP:conf/icml/MirmanGV18, DBLP:conf/ccs/DuJSZLSFYB021}.

\subsubsection{Zonotope} 
A Zonotope $\mathcal{Z}$ that abstracts $n$ variables consists of a center $\boldsymbol{\alpha}_0$ and $N$ error terms: 
$$
    \mathcal{Z} = \boldsymbol{\alpha}_0 + \sum_{i=1}^{N} \boldsymbol{\alpha}_i \cdot \epsilon_i,
$$
where $\boldsymbol{\alpha}_0, \boldsymbol{\alpha}_i \in \mathbb{R}^n$ and $\epsilon_i \in [-1, 1] (i = 1, 2, \dots, N)$. 

\subsubsection{Concretization}
The numerical bounds of the variables can be derived using the following equation:
\begin{equation} \label{eq:conc}
    \boldsymbol{l} = \boldsymbol{\alpha}_0 - \sum_{i=1}^{N} |\boldsymbol{\alpha}_i|, \ 
    \boldsymbol{u} = \boldsymbol{\alpha}_0 + \sum_{i=1}^{N} |\boldsymbol{\alpha}_i|,
\end{equation}
where $\boldsymbol{l} \in \mathbb{R}^n$ and $\boldsymbol{u} \in \mathbb{R}^n$ are the lower and upper bound, respectively.

\subsubsection{Affine Abstract Transformer} 
Given an affine transformation $\boldsymbol{W} \cdot \boldsymbol{x} + \boldsymbol{b}$, its abstract transformer for $\boldsymbol{W} \cdot \mathcal{Z} + \boldsymbol{b}$ is defined as follows:
\begin{equation} \label{eq:aff}
    \boldsymbol{W} \cdot \mathcal{Z} + \boldsymbol{b} = \boldsymbol{W} \cdot \boldsymbol{\alpha}_0 + \boldsymbol{b} + \sum_{i=1}^{N} \boldsymbol{W} \cdot \boldsymbol{\alpha}_i \cdot \epsilon_i.
\end{equation}

\subsubsection{Non-linear Abstract Transformer} 
Abstract transformers for non-linear functions adopt relaxations to over-approximate the output.
Take the tanh function as an example.
There are different abstract transformers for the tanh function, we describe the one used in Cert-RNN, as shown in Fig. \ref{fig:tanhb}.
Given an input Zonotope $\mathcal{Z}$, the abstract transformer first computes the numerical bounds, $\boldsymbol{l}$ and $\boldsymbol{u}$, using Equation \ref{eq:conc}.
Then, it computes two parallel bounding lines that minimize their distance for each variable. 
Let $y = k_jx + c_j$ and $y = k_jx + d_j$ be the upper and lower bound line of the $j$-th variable, where $k_j = (\text{tanh}(u_j) - \text{tanh}(l_j)) / (u_j - l_j), j = 1, \dots, n$. 
The output domain of the abstract transformer is:
\begin{gather}
    \text{tanh}(\mathcal{Z}) = \boldsymbol{\alpha}_0 \odot \boldsymbol{k} + \frac{\boldsymbol{c} + \boldsymbol{d}}{2} + \sum_{i=1}^{N} \boldsymbol{\alpha}_i \odot \boldsymbol{k} \cdot \epsilon_i \notag \\
    + \sum_{j=1}^{n} \frac{c_j - d_j}{2} \cdot \boldsymbol{e}_j \cdot \epsilon_{N+j},
     \label{eq:tanhp}
\end{gather}
where $\boldsymbol{k} = (k_1, k_2, \dots, k_n)^\mathsf{T}$, $\boldsymbol{c} = (c_1, c_2, \dots, c_n)^\mathsf{T}$, $\boldsymbol{d} = (d_1, d_2, \dots, $ $d_n)^\mathsf{T}$, and $\boldsymbol{e}_i$ is the $i$-th standard basis vector.

In LSTM and GRU models, there are some unique operations that are more complex, such as the Hadamard product between a sigmoid function and a tanh function.
Since the Hadamard product depends on two variables, the corresponding abstract transformer computes two planes instead of lines.
The rest calculation is similar to that in the tanh abstract transformer.
Due to the limited space, we omit the computational details, please refer to \cite{DBLP:conf/ccs/DuJSZLSFYB021}.

\subsubsection{Certification and Certified Training} 
To certify the target model's robustness in a perturbation space, certification methods first abstract the perturbation space as a Zonotope.
Then, they propagate the Zonotope through the model using abstract transformers.
Thus, they obtain an output Zonotope that is an over-approximation of the model's all possible outputs. 
Finally, let $y_t$ be the output of the correct class and $y_f$ be the incorrect class's, the robustness is proven if the lower bound of $y_t - y_f$ is positive.
Furthermore, certification methods can be extended as certified training methods by training models with a robustness loss $l$ computed according to the output Zonotope $\mathcal{Z}_o$.
For example, $l = \max_{\boldsymbol{y} \in \mathcal{Z}_o} l_{ce}(\boldsymbol{y}, l_t)$, where $l_{ce}$ is the Cross-Entropy loss and $l_t$ denotes the label of the correct class.
However, certified training methods that improve the model's robustness also cause a decrease in its clean accuracy.
To mitigate this problem, existing works \cite{DBLP:conf/iclr/ZhangCXGSLBH20} usually use a linear combination of the robustness loss and the standard loss.

\section{Motivation}
In this section, we demonstrate the vulnerability in existing certified defenses \cite{DBLP:conf/icml/KoLWDWL19,DBLP:conf/ccs/DuJSZLSFYB021} by proving that defending against multi-frame attacks is more difficult than defending against one-frame attacks.

\subsection{Defense Strategy} \label{sec:paf}

Let $\mathcal{S}^{(t)} = \{ \boldsymbol{x}^{(t)} : \Vert \boldsymbol{x}^{(t)} - \boldsymbol{x}_c^{(t)} \Vert_p \le \varepsilon \}$ be the perturbation space (i.e., the neighborhood contains all potential adversarial examples) of the $t$-th frame.

Previous works focus on one-frame attacks and thus adopt \textit{the perturb-one-frame strategy}, which considers the perturbation space of the one-frame adversary:
\begin{gather*}
    \mathcal{S}_1 = \bigcup_{t=1}^{T} \mathcal{S}_1^{(t)} = \bigcup_{t=1}^{T} \{ \boldsymbol{X} : \boldsymbol{x}^{(1)} = \boldsymbol{x}_c^{(1)}, \dots, \boldsymbol{x}^{(t-1)} = \boldsymbol{x}_c^{(t-1)}, \\
    \boldsymbol{x}^{(t)} \in \mathcal{S}^{(t)}, \boldsymbol{x}^{(t+1)} = \boldsymbol{x}_c^{(t+1)}, \dots, \boldsymbol{x}^{(T)} = \boldsymbol{x}_c^{(T)} \}.
\end{gather*}

Aiming at defending against multi-frame attacks, we adopt \textit{the perturb-all-frame strategy}, which considers the perturbation space of the worst-case multi-frame adversary:
\begin{gather*}
    \mathcal{S}_2 = \{ \boldsymbol{X} : \boldsymbol{x}^{(1)} \in \mathcal{S}^{(1)}, \dots, \boldsymbol{x}^{(t)} \in \mathcal{S}^{(t)}, \dots, \boldsymbol{x}^{(T)} \in \mathcal{S}^{(T)} \}.
\end{gather*}

\begin{lemma} \label{lem:1}
    $\mathcal{S}_1$ is a proper subset of $\mathcal{S}_2$.
\end{lemma}

\renewcommand{\IEEEQED}{\IEEEQEDopen}
\begin{IEEEproof}[Proof of Lemma \ref{lem:1}]
    On the one hand, according to their definitions, $\mathcal{S}_1^{(t)}$ is a subset of $\mathcal{S}_2, t = 1, 2, \dots, T.$ Since $\mathcal{S}_1$ is the union of all $\mathcal{S}_1^{(t)}$s, $\mathcal{S}_1$ is also a subset of $\mathcal{S}_2$. 
    On the other hand, we can find an element $\boldsymbol{X}_{adv} = [ \boldsymbol{x}_{adv}^{(1)}, \boldsymbol{x}_{adv}^{(2)}, \dots, \boldsymbol{x}_{adv}^{(T)} ] $ in $\mathcal{S}_2$ but not in $\mathcal{S}_1$, where $\boldsymbol{x}_{adv}^{(t)} \in \mathcal{S}^{(t)}, \boldsymbol{x}_{adv}^{(t)} \ne \boldsymbol{x}_c^{(t)}, t = 1, 2, \dots, T.$
    In conclusion, $\mathcal{S}_1$ is a proper subset of $\mathcal{S}_2$.
\end{IEEEproof}

\begin{theorem} \label{thm:1}
    A model that is robust against multi-frame attacks around a clean sample $\boldsymbol{X}_c$ is also robust against one-frame attacks around $\boldsymbol{X}_c$, but not vice versa.
\end{theorem}

\begin{IEEEproof}[Proof of Theorem \ref{thm:1}]
    On the one hand, the model is robust against multi-frame attacks around $\boldsymbol{X}_c$ indicates that all samples in $\mathcal{S}_2$ are robust. Since $\mathcal{S}_1$ is a proper subset of $\mathcal{S}_2$, all samples in $\mathcal{S}_1$ must be robust. Thus, the model is also robust against one-frame attacks around $\boldsymbol{X}_c$. 
    On the other hand, the model is robust against one-frame attacks around $\boldsymbol{X}_c$ only indicates that all samples in $\mathcal{S}_1$ are robust. As for samples in $\mathcal{S}_2 - \mathcal{S}_1$, their robustness is uncertain. Thus, there can be adversarial examples in $\mathcal{S}_2$.
\end{IEEEproof}

\subsection{Challenges}

\begin{figure}[t]
    \centering
    \includegraphics[width=2.2in]{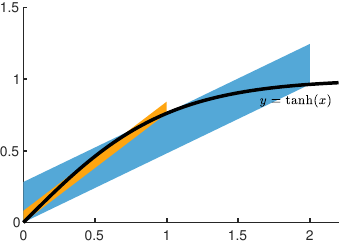}
\caption{Examples of perturbation space's effect on the linear relaxations of tanh. The orange and blue areas are the linear relaxation when the perturbation space is $[0,1]$ and $[0,2]$, respectively.}
\label{fig:relax_}
\end{figure}

While the perturb-all-frame strategy offers a stronger guarantee of models' robustness, however, it also causes a precision issue.
Most existing certified defenses apply linear relaxations to handle non-linear functions.
As shown in Fig. \ref{fig:relax_}, relaxations in smaller spaces are more precise because a shorter curve is closer to a straight line than a longer one.
As a result, larger perturbation spaces lead to less precise relaxations, which brings an intractable challenge to robustness certification against multi-frame attacks.

\section{Methodology}
In this section, we present RNN-Guard, the first certified defense against multi-frame attacks for RNN models.

\subsection{Overview}
\subsubsection{Threat Model}
We focus on white-box multi-frame adversarial attacks, which indicate the strongest adversary who has full knowledge about the target model (including its parameters and architecture) and can perturb all frames in its input sequence simultaneously by adding $\ell_p (p = \infty)$ noises to their embeddings (as we introduced in Section \ref{sec:atk}).

\subsubsection{Robustness Certification}
To certify RNN model's robustness against multi-frame attacks, we first construct an InterZono capturing all potential adversarial examples.
Then, we propagate the InterZono through the model using our abstract transformers.
The output InterZono we obtain is an over-approximation of the model's all possible outputs.
Finally, we certify the model's robustness by confirming whether $y_t - y_f$ is positive, where $y_t$ and $y_f$ are the correct and incorrect class's outputs, respectively.

\subsubsection{Certified Training}
To improve RNN model's robustness against multi-frame attacks, we first compute the output InterZono $\mathcal{D}_o$ by propagating the perturbation space through the model.
Then, we calculate a robustness loss $l = \max_{\boldsymbol{y} \in \mathcal{D}_o} l_{ce}(\boldsymbol{y}, l_t)$, where $l_{ce}$ is the Cross-Entropy loss, and $l_t$ denotes the label of the correct class.
We combine the robustness loss with the standard loss to improve both the robustness and clean accuracy of the model.
Finally, we update the model's parameters through training according to the combined loss.

\subsection{InterZono}

\subsubsection{InterZono}
We present a novel abstract domain called InterZono to address the precision issue caused by the perturb-all-frame strategy.

\begin{definition}[InterZono]
    An InterZono $\mathcal{D}$ is an intersection between two Zonotopes, which are the main domain $\mathcal{Z}_m$ and the support domain $\mathcal{Z}_s$:
$$
    \mathcal{D} =  \mathcal{Z}_m \cap \mathcal{Z}_s .
$$
\end{definition}

The main domain propagates the relations of variables through the model and the support domain is designed to refine the main domain for improving the precision.

\subsubsection{Concretization}
Given an InterZono $\mathcal{D}$, we first calculate the lower and upper bound for $\mathcal{Z}_m$ (i.e., $\boldsymbol{l}_m$ and $\boldsymbol{u}_m$) and $\mathcal{Z}_s$ (i.e., $\boldsymbol{l}_s$ and $\boldsymbol{u}_s$) according to Equation \ref{eq:conc}.
Then, the lower and upper bound of $\mathcal{D}$, i.e., $\boldsymbol{l}$ and $\boldsymbol{u}$, are computed as follows:
\begin{equation} \label{eq:bod}
	\boldsymbol{l} = \max\{ \boldsymbol{l}_m, \boldsymbol{l}_s \}, \ \boldsymbol{u} = \min \{ \boldsymbol{u}_m, \boldsymbol{u}_s \},
\end{equation}
where the maximum and minimum are taken in an element-wise manner.

\subsubsection{Affine Abstract Transformer}
The affine abstract transformer of InterZono is simple.
We apply the affine abstract transformer of Zonotope to the main domain and the support domain according to the Equation \ref{eq:aff}.
The outputs of the main domain and the support domain are the main domain and the support domain of the new InterZono, respectively.

\begin{figure*}[pt]
    \centering
    \subfloat[]{
    \includegraphics[width=1.6in]{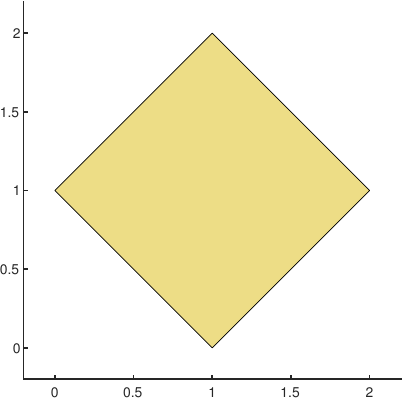}
    }
    \hfil
    \subfloat[]{
    \includegraphics[width=1.6in]{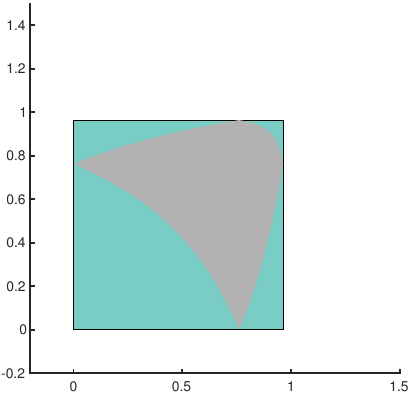}
    }
    \hfil
    \subfloat[]{
    \includegraphics[width=1.6in]{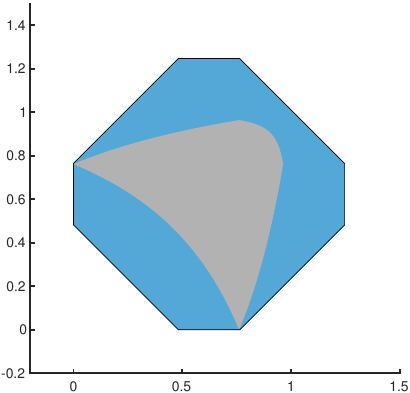}
    }
    \hfil
    \subfloat[]{
    \includegraphics[width=1.6in]{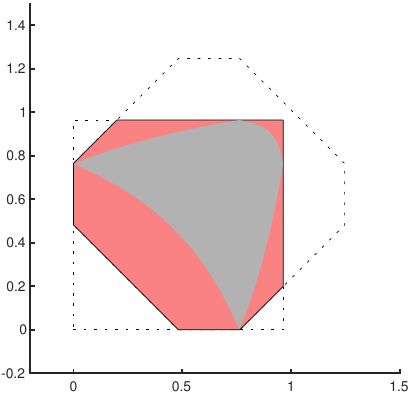}
    }
\caption{An example of different relaxations for tanh. The grey area is the exact image of the input domain under tanh. (a) Input domain. (b) Box-style relaxation's output domain. (c) Parallelogram-style relaxation's output domain. (d) Ours.}
\label{fig:exa1}
\end{figure*}

\subsubsection{Non-linear Abstract Transformer}
Unlike the affine abstract transformer, both the main domain and the support domain of the new InterZono are computed mainly based on the main domain of the input InterZono.
We continue to use the tanh abstract transformer as an example.
Given an InterZono $\mathcal{D}$, we first calculate its lower and upper bound $\boldsymbol{l}$ and $\boldsymbol{u}$ according to Equation \ref{eq:bod}.
Then, we apply the parallelogram-style abstract transformer to the main domain $\mathcal{Z}_m$, according to Equation \ref{eq:tanhp}.
The output zonotope is denoted by $\mathcal{Z}_m^\prime$.
Note that we compute the $\boldsymbol{k}$ in Equation \ref{eq:tanhp} according to the bounds of $\mathcal{D}$ rather than $\mathcal{Z}_m$.
Next, we compute the output's support domain $\mathcal{Z}_s^\prime$ as follows (i.e., the box-style abstract transformer)
\begin{equation} \label{eq:spt-dom}
    \mathcal{Z}_s^\prime = \frac{\boldsymbol{u} + \boldsymbol{l}}{2} + \sum_{i=1}^{n} \frac{u_i - l_i}{2} \cdot \boldsymbol{e}_i \cdot \epsilon_i,
\end{equation}
where $u_i$ and $l_i$ are the $i$-th component of $\boldsymbol{l}$ and $\boldsymbol{u}$, respectively.
Finally, the output InterZono $\mathcal{D}^\prime$ is $\mathcal{Z}_m^\prime \cap \mathcal{Z}_s^\prime$.

For complex operation in LSTM and GRU models, e.g., the Hadamard product between a sigmoid function and a tanh function $\sigma(x) \odot \text{tanh}(y)$, our abstract transformer is constructed in a  similar way.
Let $\mathcal{D}_x = \mathcal{Z}_{x,m} \cap \mathcal{Z}_{x,s}$ and $\mathcal{D}_y =  \mathcal{Z}_{y,m} \cap \mathcal{Z}_{y,s}$ be two InterZonos.
To calculate the output of $\sigma(\mathcal{D}_x) \odot \text{tanh}(\mathcal{D}_y)$, denoted by $\mathcal{D}_z$, we first calculate the lower and upper bounds of $\mathcal{D}_x$ and $\mathcal{D}_y$, denoted by $\boldsymbol{l}_x, \boldsymbol{u}_x$ and $\boldsymbol{l}_y, \boldsymbol{u}_y$, respectively.
Then, we apply the abstract transformer for $\sigma(x) \odot \text{tanh}(y)$ in Cert-RNN to $\mathcal{Z}_{x,m}$ and $\mathcal{Z}_{y,m}$, but replace the lower and upper bounds of $\mathcal{Z}_{x,m}$ and $\mathcal{Z}_{y,m}$ with the corresponding bounds of $\mathcal{D}_x$ and $\mathcal{D}_y$.
Let $\mathcal{Z}_{z,m}$ denote $\sigma(\mathcal{Z}_{x,m}) \odot \text{tanh}(\mathcal{Z}_{y,m})$. 
Next, we calculate the support domain of $\mathcal{D}_z$, denoted by $\mathcal{Z}_{z,s}$, as follows.
\begin{gather*}
    \mathcal{Z}_{z,s} = \frac{\boldsymbol{u} + \boldsymbol{l}}{2} + \sum_{i=1}^{n} \frac{u_i - l_i}{2} \cdot \boldsymbol{e}_i \cdot \epsilon_i, \\
    \boldsymbol{l} = \min \{ \sigma(\boldsymbol{l}_x) \odot \text{tanh}(\boldsymbol{l}_y), \sigma(\boldsymbol{u}_x) \odot \text{tanh}(\boldsymbol{l}_y) \}, \\
    \boldsymbol{u} = \max \{ \sigma(\boldsymbol{l}_x) \odot \text{tanh}(\boldsymbol{u}_y), \sigma(\boldsymbol{u}_x) \odot \text{tanh}(\boldsymbol{u}_y) \},
\end{gather*}
where the maximum and minimum are taken in an element-wise manner and $\boldsymbol{l}, \boldsymbol{u} \in \mathbb{R}^n$.
Finally, the output InterZono $\mathcal{D}_z$ is $\mathcal{Z}_{z,m} \cap \mathcal{Z}_{z,s}$.

\subsubsection{Tightness}
We prove that InterZono is tighter than Zonotope.

\begin{theorem} \label{thm:2}
    The output InterZono $\mathcal{D}^\prime$ is a tighter over-approximation (super-set) of $\mathcal{Z}_m$'s image under tanh.
\end{theorem}

\begin{IEEEproof}[Proof of Theorem \ref{thm:2}]
    Let $\mathcal{G}$ be the exact image of $\mathcal{Z}_m$ under the tanh function. 
    Since the parallelogram-style abstract transformer calculates the over-approximation of $\mathcal{Z}_m$, $\mathcal{G}$ is a subset of $\mathcal{Z}_m^\prime$.
    Meanwhile, since the box-style abstract transformer can be considered as a special case of parallelogram-style abstract transformer ($\boldsymbol{k} = 0$), $\mathcal{G}$ is also a subset of $\mathcal{Z}_s^\prime$.
    Thus, $\mathcal{G}$ is a subset of $\mathcal{D}^\prime$.
    Moreover, since $\mathcal{D}^\prime$ is the intersection between $\mathcal{Z}_m^\prime$ and $\mathcal{Z}_s^\prime$, $\mathcal{D}^\prime$ is tighter than any of them.
    This conclusion also holds for other activation function such as sigmoid.
\end{IEEEproof}

We provide an example in Fig. \ref{fig:exa1} to support Theorem \ref{thm:2}.

\subsubsection{Efficiency}
We prove that InterZono is as efficient as Zonotope.

\begin{theorem} \label{thm:3}
    The time complexity of InterZono's abstract transformers are equal to those of Zonotope's.
\end{theorem}

\begin{IEEEproof}[Proof of Theorem \ref{thm:3}]
    Let $T_Z(n) = O(f_Z(n))$ be the time complexity of Zonotope's linear abstract transformer. 
    According the definition of InterZono's linear abstract transformer, its time complexity is $T_D(n) = O(f_Z(n)) + O(f_Z(n)) = O(f_Z(n)) = T_Z(n)$.
    Let $T_Z^\prime(n) = O(f_Z^\prime(n))$ be the time complexity of Zonotope's non-linear abstract transformer.
    According the definition of InterZono's non-linear abstract transformer, the time complexity of calculating $\mathcal{Z}_m^\prime$, denoted as $T_m^\prime(n)$, is equal to $O(f_Z^\prime(n))$.
    Since the box-style abstract transformer is a special case of parallelogram-style abstract transformer without calculating and multiplying $\boldsymbol{k}$, the time complexity of calculating $\mathcal{Z}_s^\prime$, denoted as $T_s^\prime(n)$, is no larger than $O(f_Z^\prime(n))$.
    Thus, the time complexity of InterZono's non-linear abstract transformer is $T_D^\prime(n) = T_m^\prime(n) + T_s^\prime(n) = O(f_Z^\prime(n)) = T_Z^\prime(n).$
\end{IEEEproof}

\section{Evaluation}
In this section, we evaluate the performance of RNN-Guard on robustness certification against multi-frame attacks.
To the best of our knowledge, RNN-Guard is the first robustness certification method against multi-frame attacks.
Since there is no existing baseline method, we use RNN-Guard with the classic abstract domian Zonotope as the baseline method.

\subsection{Experimental Settings} \label{ss:es}

\subsubsection{Models}
We use LSTM models and GRU models, which consist 32 and 64 hidden neurons (consistent with those used in previous work \cite{DBLP:conf/ccs/DuJSZLSFYB021}).
We refer to them as LSTM-32, LSTM-64, GRU-32, and GRU-64 in the following.

\subsubsection{Datasets}
We use two datasets in our evaluation.

\noindent (\romannumeral1)
\textit{Rotten Tomatoes Movie Review (RT)} dataset \cite{DBLP:conf/acl/PangL05}, which is a benchmark corpus of movie reviews used for sentiment analysis.
The RT dataset contains about 39000 and 4800 samples in the training set and the testing set, respectively.

\noindent (\romannumeral2)
\textit{Yelp Reviews Polarity (Yelp)} dataset \cite{DBLP:conf/nips/ZhangZL15}, which is a large text classification benchmark.
The Yelp dataset contains about 560000 and 38000 samples in the training set and the testing set, respectively.

\subsubsection{Hardware}
All experiments are conducted on a server with an Intel Core i9-10920X CPU running at 3.50 GHz, 128 GB memory, 4TB HDD, and a GeForce RTX 3090 GPU card.

\subsubsection{Word Embedding}
We use the \textit{GloVe} model \cite{DBLP:conf/emnlp/PenningtonSM14} to map the words into embeddings and normalize the word embeddings to reduce their internal covariate shift.
For the out-of-vocabulary words, we randomly sample from the uniform distribution in $[-0.1, 0.1]$ for initialization.
On vanilla RNN models, we use the pre-trained model `glove.840B.300d'.
On LSTM and GRU models, due to the limited GPU memory, we use the smaller model `glove.6B.50d'.

\subsubsection{Evaluation Metrics}
We consider the following metrics:

\noindent (\romannumeral1) the \textit{certified robust accuracy} (C.Acc.) at $\varepsilon_e$ on a dataset is the fraction of samples that are guaranteed (by robustness certification) to be robust within a neighborhood of theirs with radius equal to $\varepsilon_e$ in the test set, which can be seen as the lower bound of the target model's robustness;

\noindent (\romannumeral2) the \textit{running time} of a method on a dataset is running time of the robustness certification method to certify the robustness of all samples in the testing set.

\subsection{Results and Analysis}

We compare RNN-Guard with InterZono to RNN-Guard with Zonotope across four models and two datasets to demonstrate the advantages of InterZono.

First, we compare the precision of Zonotope and InterZono. 
As we mentioned earlier, most certified defenses adopt linear relaxations to handle non-linear functions in the model, which causes the precision issue that eventually leads them to mistakenly certify that certain robust samples are non-robust.
Thus, improving the precision of robustness certification is one of the most important research directions in this area.
The precision of a robustness certification method is usually measured by the certified robust accuracy it computes.
The more precise a method is, the more samples it can certify the robustness of, thus the higher certified robust accuracy it computes.

\begin{table}[!t]
    \centering
    \caption{Certified robust accuracy and running time (in seconds) of using Zonotope and InterZono on the two datasets.}
    \begin{tabular}{|l|l|l|r|r|r|r|}
        \hline
        \multirow{2}{*}{ } & \multirow{2}{*}{Model} & \multirow{2}{*}{Acc.} & \multicolumn{2}{c|}{Zonotope} & \multicolumn{2}{c|}{InterZono} \\
        \cline{4-5} \cline{6-7}
        & & & C.Acc. & Time & C.Acc. & Time \\
        \hline
        \multirow{4}{*}{RT}
        & LSTM-32 & 73.78\% & 6.80\% & 73.8 & \textbf{14.81}\% & 73.7 \\
        & LSTM-64 & 73.63\% & 5.79\% & 73.5 & \textbf{10.58}\% & 73.6 \\
        &  GRU-32 & 73.49\% & 17.38\% & 73.5 & \textbf{17.83}\% & 73.6 \\
        &  GRU-64 & 74.71\% & 17.14\% & 74.5 & \textbf{18.18}\% & 74.5 \\
        \hline
        \multirow{4}{*}{Yelp}
        & LSTM-32 & 72.39\% & 15.67\% & 9.1e2 & \textbf{27.37}\% & 1.1e3 \\
        & LSTM-64 & 73.16\% & 14.32\% & 9.8e2 & \textbf{22.60}\% & 1.1e3 \\
        &  GRU-32 & 71.34\% & 21.69\% & 4.5e2 & \textbf{22.07}\% & 5.3e2 \\
        &  GRU-64 & 73.14\% & 28.81\% & 5.4e2 & \textbf{29.44}\% & 6.4e2 \\
        \hline
    \end{tabular}
    \label{tab:1}
\end{table}

\begin{figure}[pt]
    \centering
    \subfloat[]{
    \includegraphics[width=1.56in]{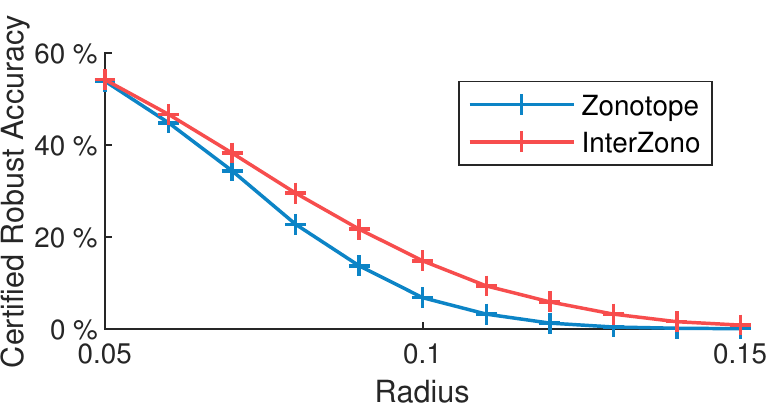}
    }
    \hfil
    \subfloat[]{
    \includegraphics[width=1.56in]{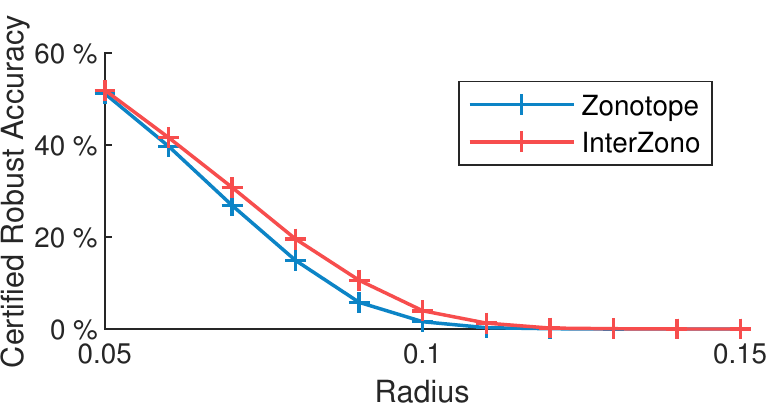}
    }
    \\
    \subfloat[]{
    \includegraphics[width=1.56in]{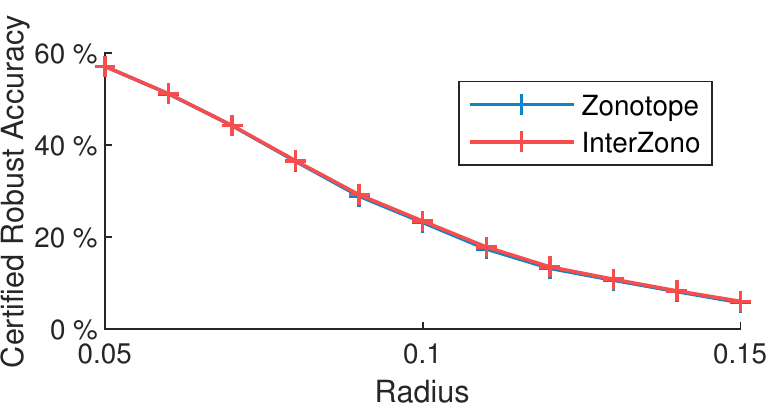}
    }
    \hfil
    \subfloat[]{
    \includegraphics[width=1.56in]{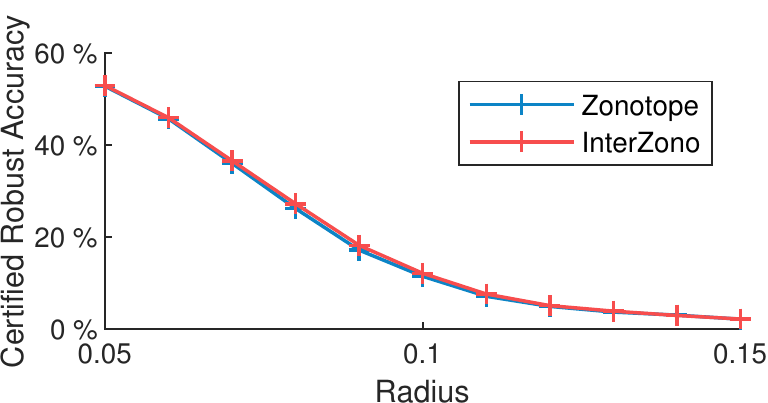}
    }
    \\
    \subfloat[]{
    \includegraphics[width=1.56in]{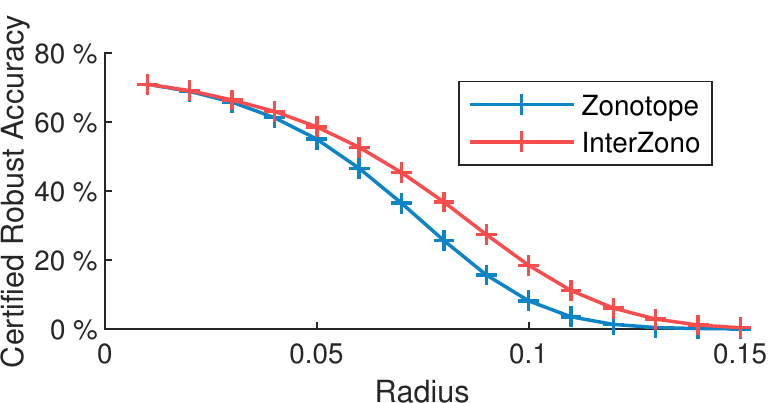}
    }
    \hfil
    \subfloat[]{
    \includegraphics[width=1.56in]{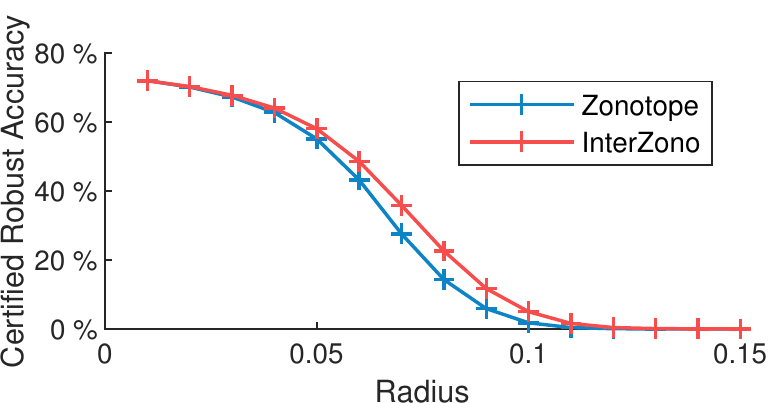}
    }
    \\
    \subfloat[]{
    \includegraphics[width=1.56in]{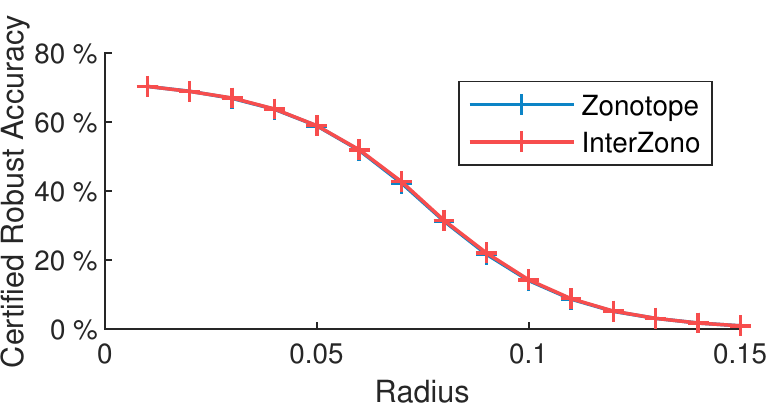}
    }
    \hfil
    \subfloat[]{
    \includegraphics[width=1.56in]{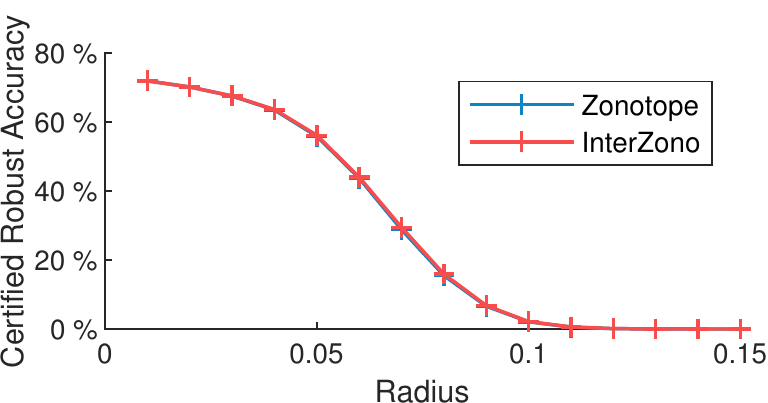}
    }
    \\
\caption{Certified robust accuracy of the four models on the RT and Yelp dataset at different $\varepsilon_e$ computed using Zonotope and InterZono. (a) RT LSTM-32. (b) RT LSTM-64. (c) RT GRU-32. (d) RT GRU-64. (e) Yelp LSTM-32. (f) Yelp LSTM-64. (g) Yelp GRU-32. (h) Yelp GRU-64.}
\label{fig:1}
\end{figure}

As shown in Table. \ref{tab:1} and Fig. \ref{fig:1}, certified robust accuracy computed using InterZono is higher than that computed using Zonotope.
For example, for the LSTM-32 model on the RT dataset, RNN-Guard with Zonotope proves that 6.80\% samples are robust while RNN-Guard with InterZono proves that 14.81\% samples are robust.
The reason for advantage of InterZono is that InterZono uses an additional support domain to refine the bounds.
Thus, InterZono is more precise than Zonotope.

Moreover, we observe that the precision improvement of InterZono is more obvious in LSTM models than in GRU models.
This is because LSTM models involve more variables and use more non-linear functions and our advantage comes from the relaxation of non-linear functions.
Hence, our method is more precise, especially on complex models with more non-linear functions.

Then, we compare the efficiency of Zonotope and InterZono. 
As shown in Table. \ref{tab:1}, the running time of RNN-Guard with InterZono is very close to those with Zonotope.
For example, for the LSTM-64 model on the RT dataset, RNN-Guard with Zonotope takes 73.5 seconds to certify the robustness of all samples in the testing set while RNN-Guard with InterZono takes 73.6 seconds.
This can be explained by Theorem \ref{thm:3}.
Thus, InterZono is as efficient as Zonotope.

In conclusion, we believe that InterZono is a better choice for certifying the robustness of RNN models.

\section{Application}
Robustness certification has many important applications, such as certifying the effectiveness of different defenses \cite{DBLP:conf/sp/GehrMDTCV18} and designing new defenses by improving the certified robustness of the target model (i.e., certified training) \cite{DBLP:conf/icml/MirmanGV18}.
In this section, we apply RNN-Guard in the above applications to further demonstrate its practicality and benefits.

\begin{figure*}[pt]
    \centering
    \subfloat[]{
    \includegraphics[width=1.62in]{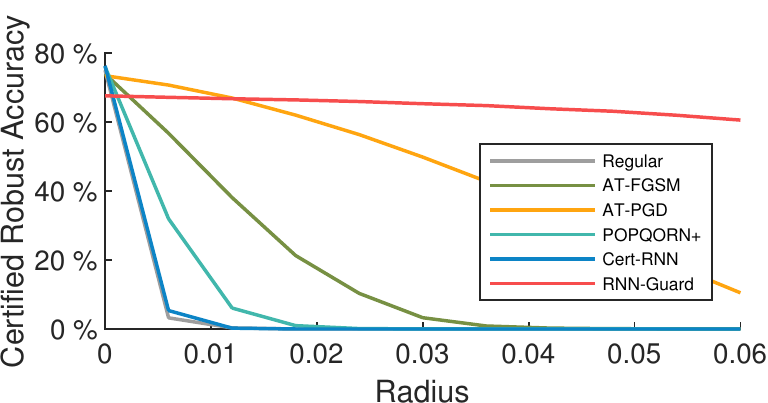}
    } \hfil
    \subfloat[]{
    \includegraphics[width=1.62in]{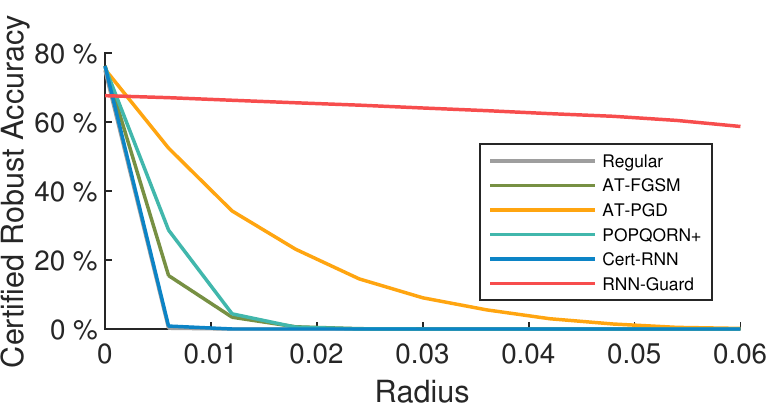}
    } \hfil
    \subfloat[]{
    \includegraphics[width=1.62in]{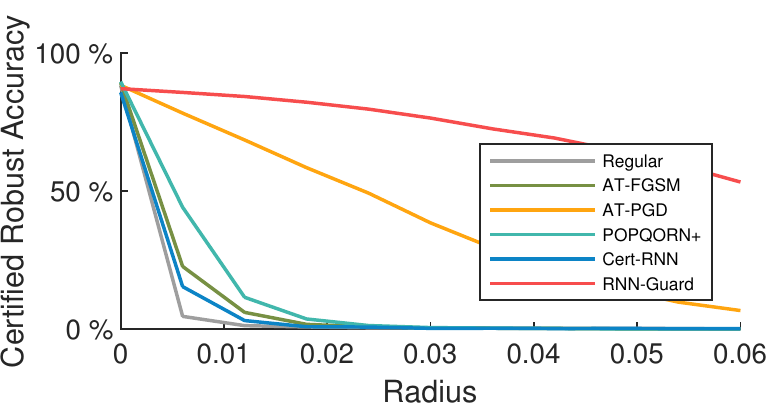}
    } \hfil
    \subfloat[]{
    \includegraphics[width=1.62in]{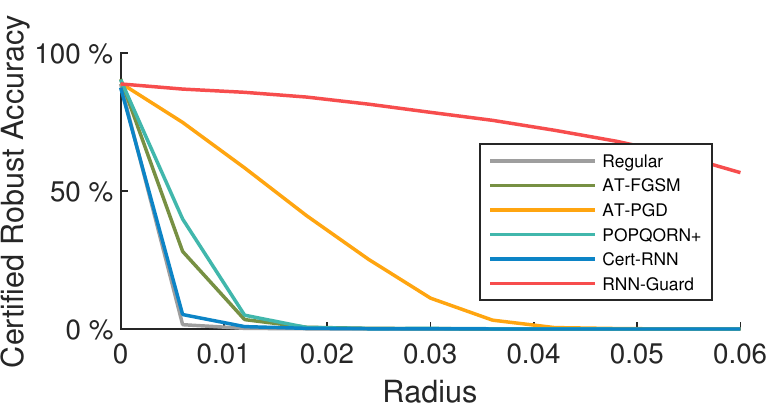}
    }
    \\
    \subfloat[]{
    \includegraphics[width=1.62in]{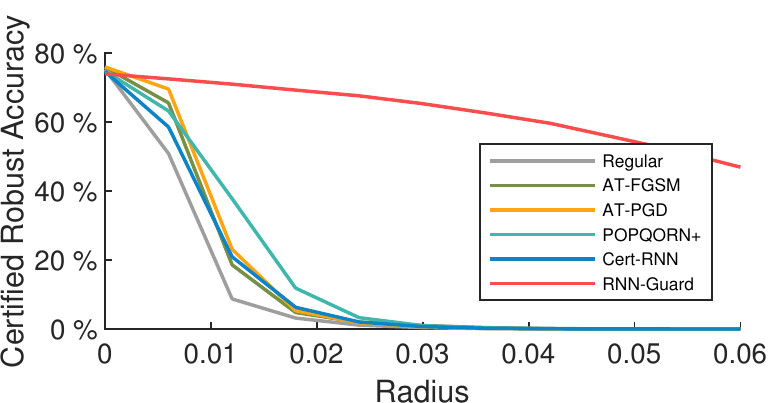}
    } \hfil
    \subfloat[]{
    \includegraphics[width=1.62in]{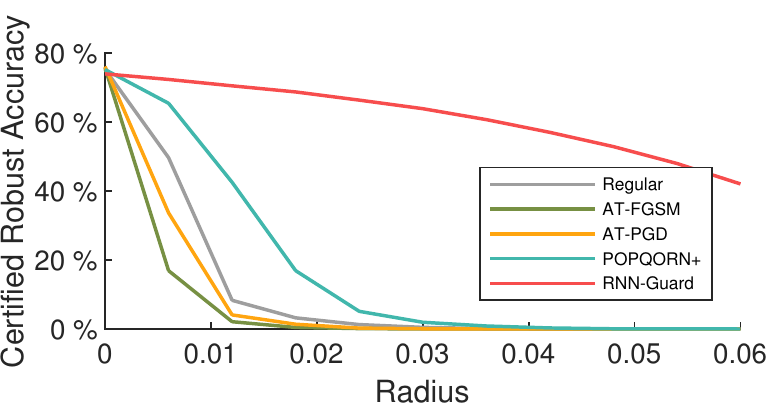}
    } \hfil
    \subfloat[]{
    \includegraphics[width=1.62in]{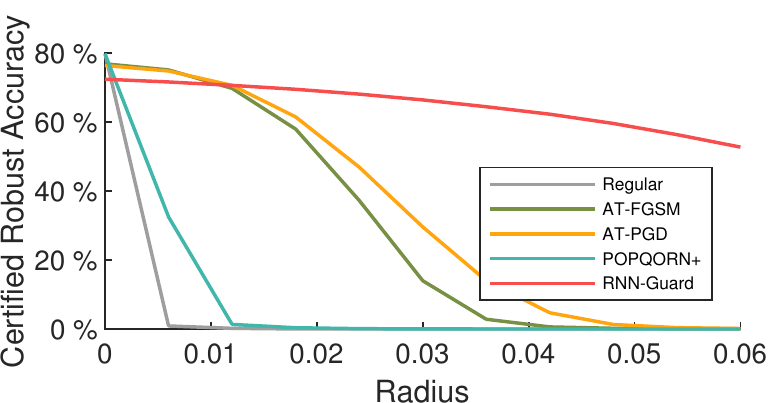}
    } \hfil
    \subfloat[]{
    \includegraphics[width=1.62in]{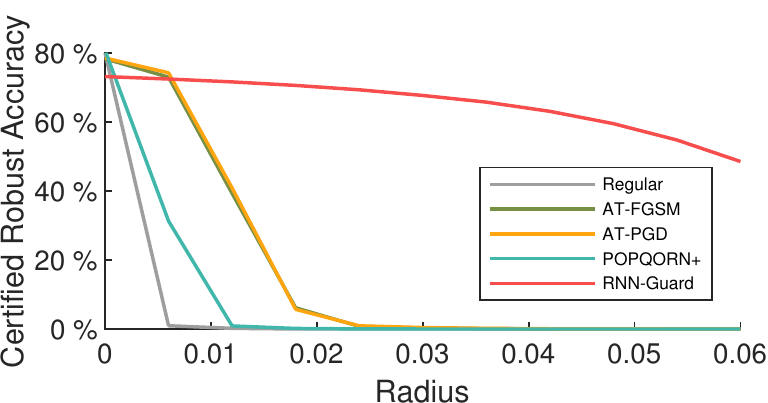}
    }
    \\
    \subfloat[]{
    \includegraphics[width=1.62in]{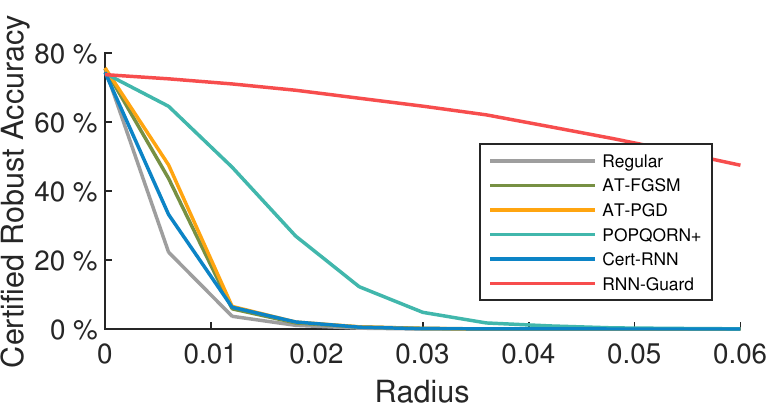}
    } \hfil
    \subfloat[]{
    \includegraphics[width=1.62in]{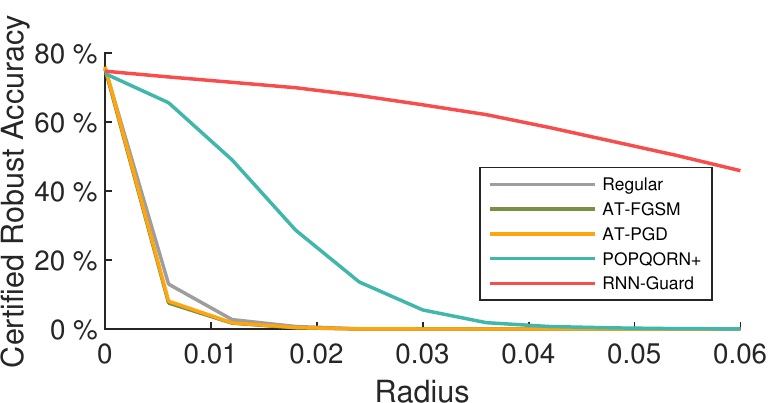}
    } \hfil
    \subfloat[]{
    \includegraphics[width=1.62in]{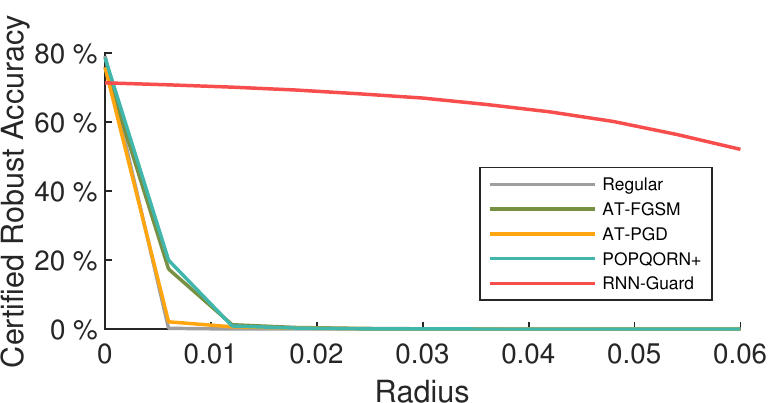}
    } \hfil
    \subfloat[]{
    \includegraphics[width=1.62in]{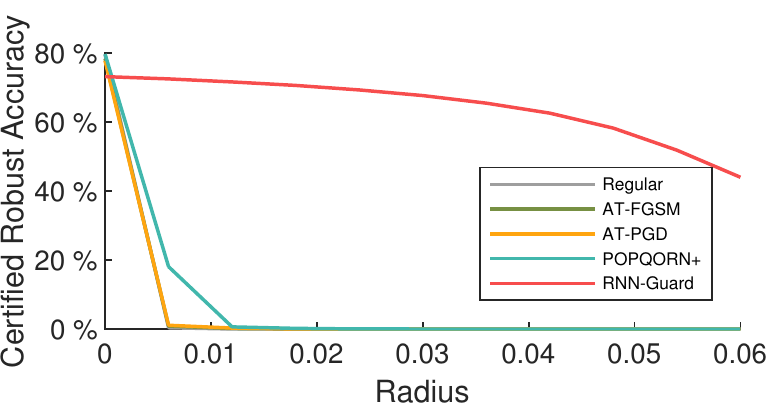}
    }
    \\
\caption{Certified robust accuracy of RNN models. On the RT dataset, we train two models with $\varepsilon_t = 0.03$ for each defense and record their certified robust accuracy in $0 \le \varepsilon_e \le 0.06$. On the TC dataset, we train two models with $\varepsilon_t = 0.05$ for each defense and record their certified robust accuracy in $0 \le \varepsilon_e \le 0.1$. (a) RT RNN-32. (b) RT RNN-64. (c) TC RNN-32. (d) TC RNN-64. (e) RT LSTM-32. (f) RT LSTM-64. (g) Yelp LSTM-32. (h) Yelp LSTM-64. (i) RT GRU-32. (j) RT GRU-64. (k) Yelp GRU-32. (l) Yelp GRU-64.}
\label{fig:2.1}
\end{figure*}

\begin{table*}[pt]
    \centering
    \caption{Certified robust accuracy of RNN models trained by the six methods. On the RT and the Yelp dataset, we set $\varepsilon_t = \varepsilon_e = 0.03$. On the TC dataset, we set $\varepsilon_t = \varepsilon_e = 0.05$. Execution failures are indicated by `-'.}
    \label{tab:2.1}
    \begin{tabular}{|l|rr|rr|rr|rr|rr|rr|}
    \hline
    \multirow{2}{*}{Method} & \multicolumn{6}{c|}{RT} & \multicolumn{2}{c|}{TC} & \multicolumn{4}{c|}{Yelp} \\
    \cline{2-7} \cline{8-9} \cline{10-13} 
    & R-32 & R-64 & L-32 & L-64 & G-32 & G-64 
    & R-32 & R-64 & L-32 & L-64 & G-32 & G-64  \\
    \hline
    Regular   
    &  0.00\% &   0.00\% &    0.41\% &  0.47\% &      0.00\% &  0.00\% 
                                                                            &  0.31\% &  0.00\% &    0.00\% &  0.00\% &      0.00\% &  0.00\% \\
    
    AT-FGSM   
    &  3.27\% &   0.00\% &    1.00\% &  0.00\% &      0.20\% &  0.00\% 
                                                                            &  0.39\% &  0.12\% &   13.99\% &  0.28\% &      0.05\% &  0.00\% \\
    
    AT-PGD    
    & 49.82\% &   9.04\% &    0.65\% &  0.06\% &      0.12\% &  0.00\% 
                                                                            & 38.44\% & 11.17\% &   29.58\% &  0.42\% &      0.01\% &  0.00\% \\
    
    POPQORN+  
    &  0.02\% &   0.00\% &    0.97\% &  1.95\% &      4.83\% &  5.52\% 
                                                                            &  0.51\% &  0.20\% &    0.01\% &  0.00\% &      0.01\% &  0.00\% \\
    
    Cert-RNN  
    &  0.00\% &   0.00\% &    0.69\% &    -    &      0.08\% &    -     
                                                                            &  0.23\% &  0.00\% &      -    &    -    &        -    &    -    \\
    
    RNN-Guard 
    & \textbf{65.29}\% & \textbf{64.05}\% &   \textbf{65.29}\% & \textbf{63.85}\% &   \textbf{64.58}\% & \textbf{64.97}\%
    
    & \textbf{76.41}\% & \textbf{78.48}\% &   \textbf{66.39}\% & \textbf{67.68}\% &   \textbf{66.93}\% & \textbf{67.65}\% \\
    \hline
    \end{tabular}
\end{table*}

\subsection{Certifying Existing Defenses} \label{ss:ced}
In this subsection, we certify the effectiveness of existing defenses, including the \textit{standard training} without any defense, two \textit{adversarial training} methods (i.e., AT-FGSM \cite{DBLP:journals/corr/GoodfellowSS14} and AT-PGD \cite{DBLP:conf/iclr/MadryMSTV18}), and two \textit{certified training} methods (i.e., POPQORN+ \cite{DBLP:conf/icml/KoLWDWL19} and Cert-RNN \cite{DBLP:conf/ccs/DuJSZLSFYB021}).
Note that Cert-RNN failed to train several large models due to timeout or running out of GPU memory.

\subsubsection{Experimental Settings}
\textit{Models.}
We consider three different kinds of RNN models, i.e., vanilla RNN models, LSTM models, and GRU models.
We will refer to them as RNN-32 (R-32), RNN-64 (R-64), LSTM-32 (L-32), LSTM-64 (L-64), GRU-32 (G-32), and GRU-64 (G-64) in the following.
\textit{Dataset.}
In addition to the RT and Yelp datasets, we use the \textit{Toxic Comment (TC)} dataset \cite{ToxicComment} provided by Kaggle, which is also used in the evaluation of Cert-RNN.
To be consistent with Cert-RNN, we also perform binary classification on this dataset by reclassifying the six toxicity categories as one.
We randomly sample a balanced subset of this dataset for evaluation.
\textit{Implementation Details.}
Though POPQORN \cite{DBLP:conf/icml/KoLWDWL19} is proposed to quantify the robustness of RNN models, we modify it for certified training and refer to it as `POPQORN+'.
The rest settings are consistent with those in Section \ref{ss:es}.

\subsubsection{Results and Analysis}
The results are shown in Tab. \ref{tab:2.1} and Fig. \ref{fig:2.1}, from which we have the following observations.

First, the standard training method achieves nearly zero certified robust accuracy in Tab. \ref{tab:2.1}.
This indicates that models trained by it lack robustness, which is already widely acknowledged by the security community.
Second, adversarial training methods achieve higher but limited certified robust accuracy than the standard training method.
For example, AT-FGSM and AP-PGD achieve 13.99\% and 29.58\% certified robust accuracy at $\varepsilon_e = 0.03$ for the LSTM-32 model on the Yelp dataset, respectively.
The reason for such limited robustness is that they are based on heuristics and thus lack theoretical guarantees.
Finally, existing certified training methods achieve relatively low certified robust accuracy.
For instance, POPQORN+ and Cert-RNN achieve 0.97\% and 0.69\% certified robust accuracy at $\varepsilon_e = 0.03$ for the LSTM-32 model on the RT dataset, respectively.
The reason for such low robustness is that they focus on one-frame attacks and defending against multi-frame attacks is more difficult.

In conclusion, multi-frame attacks pose a significant challenge to the effectiveness of existing defenses.

\subsection{Improving RNN's Robustness}
To address the security challenge posed by multi-frame attacks, we extend our certification method, i.e., RNN-Guard, as a \textit{certified training} method and compare it to existing defenses to demonstrate its advantages.

\subsubsection{Experimental Settings}
We consider the following metrics:
(\romannumeral1) the \textit{certified robust accuracy}, which is explained in Section \ref{ss:es};
(\romannumeral2) the \textit{empirical robust accuracy} (E.Acc.) at $\varepsilon_e$ on a dataset is the fraction of samples that adversarial attackers
failed to attack within a neighborhood of theirs with a radius equal to $\varepsilon_e$ in the test set, which can be seen as the upper bound of the target model's robustness; 
(\romannumeral3) the \textit{clean accuracy} (Acc.) on a dataset is the fraction of samples that is correctly classified in the test set; 
(\romannumeral4) the \textit{running time} of a method on a dataset is running time of the certified training method to train the target model for one epoch on the training set.
The rest settings are consistent with those in Section \ref{ss:ced}.

\begin{table*}[pt]
    \centering
    \caption{Empirical robust accuracy of RNN models trained by the six methods. On the RT and the Yelp dataset, we set $\varepsilon_t = \varepsilon_e = 0.03$. On the TC dataset, we set $\varepsilon_t = \varepsilon_e = 0.05$. Execution failures are indicated by `-'.}
    \label{tab:2.2}
    \begin{tabular}{|l|rr|rr|rr|rr|rr|rr|}
    \hline
    \multirow{2}{*}{Method} & \multicolumn{6}{c|}{RT} & \multicolumn{2}{c|}{TC} & \multicolumn{4}{c|}{Yelp} \\
    \cline{2-7} \cline{8-9} \cline{10-13} 
    & R-32 & R-64 & L-32 & L-64 & G-32 & G-64 
    & R-32 & R-64 & L-32 & L-64 & G-32 & G-64  \\
    \hline
    Regular   
    & 26.46\% & 27.21\% &    44.70\% & 48.44\% &     47.18\% & 51.83\% 
                                                                            & 12.58\% & 26.72\% &    13.37\% & 33.55\% &     13.34\% & 26.56\% \\
    
    AT-FGSM   
    & 50.89\% & 44.17\% &    62.23\% & 56.01\% &     64.87\% & 58.69\% 
                                                                            & 23.67\% & 57.50\% &    70.75\% & 71.75\% &     69.72\% & 71.59\% \\
    
    AT-PGD    
    & 67.00\% & 67.28\% &    62.14\% & 59.99\% &     64.56\% & 63.08\% 
                                                                            & 82.70\% & 81.95\% &    71.43\% & 72.47\% &     69.25\% & 72.12\% \\
    
    POPQORN+  
    &  4.45\% &  4.91\% &    38.91\% & 43.38\% &     46.32\% & 45.33\% 
                                                                            &  9.80\% & 12.15\% &    16.41\% & 21.91\% &     14.24\% & 16.22\% \\
    
    Cert-RNN  
    &  1.69\% &  1.22\% &    36.19\% &    -    &     42.71\% &    -    
                                                                            &  9.65\% & 14.10\% &      -     &    -    &        -    &    -     \\
    
    RNN-Guard 
    & 65.68\% & 65.09\% &    67.61\% & 66.43\% &     67.14\% & 68.05\% 
                                                                            & 78.55\% & 81.13\% &    68.53\% & 69.99\% &     68.64\% & 69.90\% \\
    \hline
    \end{tabular}
\end{table*}

\begin{table*}[pt]
    \centering
    \caption{Clean accuracy of RNN models trained by the six methods. Execution failures are indicated by `-'.}
    \label{tab:2.3}
    \begin{tabular}{|l|rr|rr|rr|rr|rr|rr|}
    \hline
    \multirow{2}{*}{Method} & \multicolumn{6}{c|}{RT} & \multicolumn{2}{c|}{TC} & \multicolumn{4}{c|}{Yelp} \\
    \cline{2-7} \cline{8-9} \cline{10-13} 
    & R-32 & R-64 & L-32 & L-64 & G-32 & G-64 
    & R-32 & R-64 & L-32 & L-64 & G-32 & G-64  \\
    \hline
    Regular   
    & 75.00\% & 74.95\% &    75.24\% & 75.28\% &     75.46\% & 75.58\% 
                                                                            & 88.24\% & 88.78\% &    79.83\% & 80.00\% &     79.13\% & 79.60\% \\
    
    AT-FGSM   
    & 73.92\% & 76.29\% &    75.89\% & 75.67\% &     75.50\% & 75.42\% 
                                                                            & 89.14\% & 90.54\% &    76.85\% & 78.32\% &     75.89\% & 78.37\% \\
    
    AT-PGD    
    & 73.41\% & 75.28\% &    75.95\% & 76.09\% &     75.67\% & 76.03\% 
                                                                            & 88.28\% & 89.10\% &    76.36\% & 78.53\% &     75.80\% & 77.75\% \\
    
    POPQORN+  
    & 76.01\% & 75.89\% &    74.77\% & 75.18\% &     74.10\% & 74.02\% 
                                                                            & 89.60\% & 90.27\% &    79.71\% & 80.76\% &     78.61\% & 79.68\% \\
    
    Cert-RNN  
    & 76.29\% & 76.21\% &    74.61\% &    -    &     74.37\% &    -    
                                                                            & 85.70\% & 87.42\% &       -    &    -    &        -    &    -    \\
    
    RNN-Guard 
    & 67.58\% & 67.62\% &    73.94\% & 73.94\% &     73.72\% & 74.71\% 
                                                                            & 87.07\% & 88.78\% &    72.39\% & 73.16\% &     71.34\% & 73.14\% \\
    \hline
    \end{tabular}
\end{table*}

\begin{table*}[pt]
    \centering
    \caption{Running Times of training RNN models for one epoch with the six methods. All results are in seconds. Execution failures are indicated by `-'.}
    \label{tab:2.4}
    \begin{tabular}{|l|rr|rr|rr|rr|rr|rr|}
    \hline
    \multirow{2}{*}{Method} & \multicolumn{6}{c|}{RT} & \multicolumn{2}{c|}{TC} & \multicolumn{4}{c|}{Yelp} \\
    \cline{2-7} \cline{8-9} \cline{10-13} 
    & R-32 & R-64 & L-32 & L-64 & G-32 & G-64 
    & R-32 & R-64 & L-32 & L-64 & G-32 & G-64  \\
    \hline
    Regular   
    &   3.7 &   3.8 &     9.4 &   9.4 &       8.0 &   8.1 
                                                                            &   1.3 &   1.3 &    140.1 & 140.1 &     120.1 & 120.0 \\
    
    AT-FGSM   
    &   7.7 &   7.7 &    23.2 &  23.1 &      19.5 &  19.5 
                                                                            &   2.7 &   2.7 &    364.3 & 363.4 &     303.0 & 302.6 \\
    
    AT-PGD    
    &  45.7 &  45.6 &   152.7 & 153.0 &     128.3 & 128.1
                                                                            &  17.0 &  16.2 &    2509.5& 2509.7&     2084.4& 2086.5\\
    
    POPQORN+  
    & 138.7 & 141.0 &   460.8 & 467.3 &     444.6 & 446.5
                                                                            &  51.2 &  51.7 &    8336.3& 8422.2&     8026.6& 8036.1\\
    
    Cert-RNN  
    & 245.1 & 256.1 &  14392.9&   -   &    12125.2&   -   
                                                                            &  90.1 &  94.3 &      -   &   -   &       -   &   -   \\
    
    RNN-Guard 
    &  40.7 &  57.0 &   3486.6& 3740.3&     2334.3& 2570.5
                                                                            &  14.8 &  20.9 &   45925.8&52670.0&    33451.4&37195.0\\
    \hline
    \end{tabular}
\end{table*}

\subsubsection{Results and Analysis}
The results are shown in Tab. \ref{tab:2.1}, \ref{tab:2.2}, \ref{tab:2.3}, \ref{tab:2.4}, and Fig. \ref{fig:2.1}, from which we have the following observations.

First, as shown in Tab. \ref{tab:2.1} and Fig. \ref{fig:2.1}, RNN-Guard achieves the highest certified robust accuracy across different models and datasets.
For example, RNN-Guard achieved $64.05\%$ certified robust accuracy at $\varepsilon_e = 0.03$ for the RNN-64 model on the RT dataset and the second-best method only reaches $9.04\%$ certified robust accuracy.
RNN-Guard attains better performance than adversarial training methods because it provides theoretical guarantees for RNN models' robustness.
RNN-Guard outperforms existing certified training methods because we focus on multi-frame attacks that is more difficult to defend against.
Roughly speaking, let $|\mathcal{S}^{(t)}|$ be the number of potential adversarial examples in $\mathcal{S}^{(t)}$ (see details in Section \ref{sec:paf}), which is a very large number, existing certified training methods improve the robustness of samples in $\mathcal{S}_1$, whose number is $|\mathcal{S}_1| = |\mathcal{S}^{(1)}| + |\mathcal{S}^{(2)}| + \dots + |\mathcal{S}^{(T)}|$.
Meanwhile, RNN-Guard improves the robustness of samples in $\mathcal{S}_2$ and their number is $|\mathcal{S}_2| = |\mathcal{S}^{(1)}| \times |\mathcal{S}^{(2)}| \times \dots \times |\mathcal{S}^{(T)}|$.
As we can see, $|\mathcal{S}_2|$ is far larger than $|\mathcal{S}_1|$, which explains why POPQORN+ and Cert-RNN achieve such low certified robust accuracy against multi-frame attacks.
Moreover, we further validate RNN-Guard's advantages by recording certified robust accuracy of different defenses at different radii in Fig. \ref{fig:2.1}.
As shown in Fig. \ref{fig:2.1}, RNN-Guard achieves the highest certified robust accuracy at almost all $\varepsilon_e$, which indicates that the advantages of RNN-Guard are general and do not rely on a specific radius.
For instance, the certified robust accuracy of RNN-Guard is the highest as $\varepsilon_e$ increases from $0.01$ to $0.06$ for the LSTM-32 model on the RT dataset.
Thus, RNN-Guard provides more robustness for RNN models.

Second, as shown in Tab. \ref{tab:2.2}, RNN-Guard attains high empirical robust accuracy across different models and datasets.
For example, RNN-Guard achieved $67.61\%$ empirical robust accuracy at $\varepsilon_e = 0.03$ for the LSTM-32 model on the RT dataset, which is the best result in this setting.
The outstanding effectiveness of RNN-Guard against adversarial attacks is guaranteed by its certified robustness results.
Though adversarial training methods can improve models' robustness to some extent, they are usually challenged by later and stronger attacks \cite{DBLP:conf/iclr/MadryMSTV18}.
Models trained with existing certified training methods cannot defend against the PGD attack (multi-frame), which is no surprise because defending against multi-frame attacks is more difficult.
In addition, though the empirical robust accuracy of AT-PGD is similar or even slightly higher than RNN-Guard, we argue that, due to the limitation of empirical robust accuracy, it doesn't mean that models trained by AT-PGD are more robust than those trained by RNN-Guard.
We conducted further experiments where we replace the PGD attack with adaptive attacks to support our argument, which are detailedly presented in Section \ref{ss:ae}.
Nevertheless, RNN-Guard is an effective defense against adversarial attacks.

\begin{table*}[t]
    \centering
    \caption{Adaptive robust accuracy of RNN models trained by the six methods. Execution failures are indicated by `-'.}
    \label{tab:3}
    \begin{tabular}{|l|rr|rr|rr|rr|}  
    \hline
    \multirow{2}{*}{Method} & \multicolumn{6}{c|}{RT} & \multicolumn{2}{c|}{TC} \\ 
    \cline{2-7} \cline{8-9} 
    & RNN-32 & RNN-64 & LSTM-32 & LSTM-64 & GRU-32 & GRU-64 
    & RNN-32 & RNN-64 \\ 
    \hline
    Regular   
    &  6.25\% &  5.07\% &    30.32\% & 36.59\% &     42.60\% & 44.92\% 
                                                                            &  0.07\% &  0.15\% \\ 
    
    AT-FGSM   
    & 20.47\% & 20.87\% &    54.26\% & 41.28\% &     51.78\% & 33.42\% 
                                                                            &  0.70\% &  1.32\% \\ 
    
    AT-PGD    
    & 29.54\% & 43.40\% &    46.89\% & 45.30\% &     54.02\% & 51.95\% 
                                                                            & 26.40\% & 18.35\% \\ 
    
    POPQORN+  
    &  2.21\% &  1.27\% &    30.11\% & 33.36\% &     36.88\% & 12.69\% 
                                                                            &  0.07\% &  0.46\% \\ 
    
    Cert-RNN  
    &  0.60\% &  0.32\% &    23.21\% &    -    &     34.93\% &    -    
                                                                            &  0.00\% &  0.00\% \\ 
    
    RNN-Guard 
    & \textbf{61.37}\% & \textbf{62.12}\% &    \textbf{56.80}\% & \textbf{55.89}\% &     \textbf{57.94}\% & \textbf{58.85}\% 
                                                                            & \textbf{36.56}\% & \textbf{42.96}\% \\ 
    \hline
    \end{tabular}
\end{table*}

Third, as shown in Tab. \ref{tab:2.3}, RNN-Guard maintains relatively high clean accuracy.
For instance, the clean accuracy of the LSTM-32 model trained with RNN-Guard on the RT dataset is $73.94\%$, which is only about $2\%$ lower than the model trained with the regular training.
The clean accuracy of models trained with RNN-Guard is lower than those trained with the regular training, but the decline is acceptable.
First, as many previous works (both empirical and certified defenses) \cite{DBLP:conf/iclr/MadryMSTV18,DBLP:conf/icml/MirmanGV18,DBLP:conf/iccv/GowalDSBQUAMK19,DBLP:conf/iclr/ZhangCXGSLBH20} have shown, improving a model's robustness can cause a drop in its clean accuracy due to the trade-off between the robustness and clean accuracy of the model.
Second, the gap between RNN-Guard and the regular training in clean accuracy is similar to those in previous works.
For example, adversarial training with PGD caused about $13\%$ decline in clean accuracy on the CIFAR-10 dataset \cite{DBLP:conf/iclr/MadryMSTV18}, certified training with IBP caused about $10\%$ decline in clean accuracy on the SVHN dataset \cite{DBLP:conf/iccv/GowalDSBQUAMK19}, and certified training with DiffAI(hSmooth) caused about $10\%$ decline in clean accuracy on the F-MNIST dataset \cite{DBLP:conf/icml/MirmanGV18}.
Last but not least, models with high robustness and lower clean accuracy are still useful in the real world.
For instance, recent works \cite{DBLP:conf/iclr/MullerBV21,DBLP:conf/ccs/0001M21} combine the advantages of models with high robustness but lower clean accuracy and those with high clean accuracy but lower robustness using model aggregation mechanisms to build a new model that achieves both high clean accuracy and high robustness.
Though the training model using RNN-Guard can result in a lower clean accuracy, we argue that RNN-Guard is still an outstanding defense because it achieves a remarkably high robustness against multi-frame attacks.

Finally, as shown in Tab. \ref{tab:2.4}, RNN-Guard is more efficient than Cert-RNN.
For instance, it only takes about 1 hour for RNN-Guard to train the LSTM-32 model for one epoch on the RT dataset, whereas Cert-RNN takes about 4 hours.
Thus, RNN-Guard is more efficient than the current SOTA method Cert-RNN, especially for long input sequences.
Moreover, RNN-Guard's training time is acceptable compared to certified training methods for CNNs.
For example, existing study \cite{DBLP:conf/iclr/ZhangCXGSLBH20} shows that CROWN-IBP takes 954 to 4173 seconds to train models with medium size on the MNIST and CIFAR-10 datasets while convex adversarial polytopes (CAP) takes 6961 to 160764 seconds to train same models on same datasets.
Note that both adversarial training and certified training consume extra time to compute an additional robustness loss in the training phase, which has \textit{no negative effect} on the model's efficiency in the \textit{inference phase}.

In conclusion, RNN-Guard achieve the highest certified robustness among the six training methods, the running time of RNN-Guard is significantly shorter than the SOTA certified training method Cert-RNN, and the decrease in accuracy caused by RNN-Guard is similar to those caused by previous certified defenses for CNNs.
Thus, we believe RNN-Guard is currently the best choice for enhancing the robustness of RNN models.

\subsection{Adaptive Evaluation of Existing Defenses and RNN-Guard} \label{ss:ae}
In this subsection, we further evaluate the defensive effectiveness of existing defenses and RNN-Guard using adaptive attacks, which are considered to be reliable evaluations of adversarial defenses.
We use an automatic adaptive attack tool called \textit{AdaptiveAutoAttack} (A3) \cite{NEURIPS2021_e1c13a13}, which is experimentally demonstrated to be more effective than traditional manually-designed adaptive attacks.

\subsubsection{Experimental Settings}
\textit{Automatic Adaptive Attack.}
In the above section, we empirically evaluate the defensive effectiveness of different defenses using the PGD attack.
However, such a one-sided evaluation cannot reveal the true vulnerability of each defense.
To comprehensively evaluate the robustness of models trained by different defenses against adversarial attacks, we replace the PGD attack with the adaptive attack, which can utilize the weakness in the design of the defense to adaptively generate adversarial examples for breaking through the defense.
In the normal adaptive attack setting, the adversary has \textit{complete knowledge} of the defense and manually crafts attacks based on their knowledge.
However, most common adaptive attack methods are designed for empirical defenses such as adversarial training, which are ineffective for certified defenses because common adaptive attack methods such as replacing the loss function optimized by the attack are ineffective for certified defenses.
Therefore, we follow \cite{NEURIPS2021_e1c13a13} to design our adaptive attack, which searches for an effective attack over different combinations of common reusable building blocks of existing adaptive attacks (i.e., attack algorithm and parameters, network transformations, and loss functions).
In our experiment, we use the AdaptiveAutoAttack (A3) tool with a little customization for the adaptation of the input data.

\textit{Evaluation Metrics.}
We use \textit{adaptive robust accuracy} to quantify the defense effectiveness of a defense against adaptive attackers, 
which is the fraction of samples that adaptive adversaries fail to find their adversarial examples in the test set.

The rest of the experimental settings are consistent with those in the above section, which are explained in Section \ref{ss:es} and \ref{ss:ced}.
Due to timeout, results on the Yelp dataset are not presented.

\subsubsection{Results and Analysis}

We compare the adaptive robust accuracy of different defenses, which reveals the defensive effectiveness of a defense against the worst-case adversary who has complete knowledge of that defense.
The results are shown in Tab. \ref{tab:3}, from which we have the following observations.

First, RNN-Guard achieves the highest adaptive robust accuracy across different models and datasets.
For example, RNN-Guard achieved $61.37\%$ empirical robust accuracy at $\varepsilon_e = 0.03$ for the RNN-32 model on the RT dataset, while the second-best method only reached $29.54\%$.
The advantages of RNN-Guard in certified robustness provide a solid foundation for its effectiveness against adaptive adversaries.
Hence, it is extremely difficult for adaptive adversaries to successfully attack models trained by RNN-Guard.

Second, adversarial training is vulnerable to adaptive attacks because they are empirical defenses and lack theoretical guarantees for security.
For instance, comparing the results in Tab. \ref{tab:2.2}, the RNN-32 model trained by AT-PGD on the RT dataset achieves $67.00\%$ empirical robust accuracy against the PGD attack, but it only achieves $29.54\%$ adaptive robust accuracy against the adaptive attack.
In contrast, the RNN-32 model trained by RNN-Guard on the RT dataset achieves $65.68\%$ empirical robust accuracy and $61.37\%$ adaptive robust accuracy, which indicates that RNN-Guard is a more dependable defense against the adaptive attack.

Finally, existing certified training methods cannot defend against the adaptive attack because they focus on one-frame attacks and defending against multi-frame attacks is beyond their capability.
For example, POPQORN+ and Cert-RNN achieve $2.21\%$ and $0.60\%$ adaptive robust accuracy for the RNN-32 model on the RT dataset, respectively.

In conclusion, the results of adaptive attacks further confirm that RNN-Guard is the best choice for enhancing the robustness of RNN models against multi-frame attacks.

\section{Discussion}

\textit{Evaluation on commercial models/platforms.}
With the great success of Machine Learning as a Service (MLaaS), many companies have launched their own models/platforms for NLP tasks, such as Google Cloud NLP, IBM Waston Natural Language Understanding, Microsoft Azure Text Analytics, and Amazon AWS Comprehend.
For security and copyright considerations, those companies only provide APIs to their customers without any access to their models' structure or parameters.
However, certified defenses (e.g., POPQORN, Cert-RNN, and RNN-Guard) are white-box evaluations, i.e., they require full knowledge of the model.
Besides, the attack methods we used in the evaluation (i.e., FGSM, PGD, and adaptive attacks) are white-box attacks, which also require full knowledge of the target model.
Thus, without full access to the model, it is nearly impossible to conduct any white-box evaluation on those commercial models/platforms.
Nevertheless, the large datasets and complex model structures we used in our evaluations are chosen according to those in real-world applications.
Thus, we believe our results have certain practical significance.

\textit{Extending to SOTA models.}
Recently, SOTA models for NLP tasks have shifted from RNNs to Transformers, especially large-scale pre-trained models such as BERT \cite{DBLP:conf/naacl/DevlinCLT19} and GPT-3 \cite{DBLP:conf/nips/BrownMRSKDNSSAA20}.
For instance, the base BERT model has 12 layers and 110M parameters.
Extending existing certified defense to such models is extremely hard due to their enormous number of parameters.
However, we argue that Transformers cannot completely replace RNNs because RNNs are more suitable for simple tasks due to their small size and adequate performance.
We will conduct research on extending RNN-Guard to Transformers in the future.

\textit{Extending to other tasks.}
In this work, we focus on several common textual classification tasks such as sentiment analysis and toxic content detection.
Besides classification tasks, there are many other tasks that RNNs can handle including sequence prediction, machine translation, and question answering.
However, the current definition of robustness is proposed for classification tasks, which require a ground-truth label to determine whether a sample is robust or not.
Thus, the existing definition of robustness cannot directly apply to other tasks due to the lack of ground-truth labels in them.
We will conduct research on proposing a general definition of robustness and extending RNN-Guard to general tasks in future works.

\textit{Extending to other threat models.}
In this work, we follow the threat model in the previous works \cite{DBLP:conf/ccs/DuJSZLSFYB021} where attackers can directly perturb word embeddings because word embeddings are in a continuous space and it is easy to express all potential adversarial examples with an abstract domain. 
Besides this threat model, there exist attacks on other attack surfaces such as word substitution attacks \cite{DBLP:conf/emnlp/AlzantotSEHSC18} and character injection attacks \cite{DBLP:journals/corr/abs-2106-09898}.
However, words and characters are in discrete space and how to extend abstract domains to discrete space remains a challenging problem.
We are considering extending RNN-Guard to other threat models in our future plan.

\section{Related Work}
Adversarial attacks and defenses have been one of the most popular topics in the machine learning area over the past few years.

\textit{Adversarial Attacks.}
The existence of adversarial attacks was first found on feed-forward networks such as convolutional neural networks (CNNs) \cite{DBLP:journals/corr/SzegedyZSBEGF13}.
Afterward, many studies \cite{DBLP:conf/milcom/PapernotMSH16,DBLP:conf/acl/EbrahimiRLD18,DBLP:conf/ijcai/SatoSS018} show that RNNs are also vulnerable to adversarial examples.
Some attackers perturb the input text by replacing a few characters or words \cite{DBLP:conf/acl/EbrahimiRLD18}.
Other attackers perturb the word embedding using $\mathcal{L}_p$-bounded attacks \cite{DBLP:conf/ijcai/SatoSS018}, which are considered to be more powerful \cite{DBLP:conf/ccs/DuJSZLSFYB021}.
Thus, in our threat model, we assume that adversaries can directly perturb the word embedding of textual input sequences.

\textit{Empirical Defenses.}
Early defensive works are based on heuristics to defend against adversarial attacks empirically, including \textit{adversarial training} \cite{DBLP:journals/corr/GoodfellowSS14,DBLP:conf/cvpr/ShrivastavaPTSW17}, \textit{model distillation} \cite{DBLP:conf/sp/PapernotM0JS16}, and \textit{feature denoising} \cite{DBLP:conf/cvpr/XieWMYH19}.
For example, adversarial training forces the target model to memorize adversarial examples by adding them to the training set.
However, empirical defenses lack theoretical guarantees and can be defeated by stronger attacks \cite{DBLP:conf/icml/AthalyeC018,DBLP:conf/nips/TramerCBM20}.
To end the constant competition between attackers and defenders, recent researches have started to focus on certified defenses.

\textit{Certified Defenses.}
Certified defenses, or robustness certification, aim to formally verify whether a given neighbor around the clean input contains any adversarial example, which theoretically guarantees the safety of the target models.
The existing certified defenses can fall into one of two categories, complete methods or incomplete methods. 
Complete methods usually model the robustness certification problem as a \textit{satisfiability modulo theories} (SMT) \cite{DBLP:conf/atva/Ehlers17,DBLP:conf/cav/KatzBDJK17,DBLP:conf/nips/BunelTTKM18} or \textit{mixed integer linear programming} (MILP) \cite{DBLP:journals/corr/LomuscioM17,DBLP:conf/atva/ChengNR17,DBLP:conf/iclr/TjengXT19} problem. 
Though complete methods can derive precise results of the model's robustness, they are constrained to small models with piece-wise linear activation functions.

To certify the robustness of large models with general activation functions, incomplete methods employ relaxed approaches including \textit{Linear Programming} (LP) \cite{DBLP:conf/icml/WengZCSHDBD18,DBLP:conf/nips/SalmanY0HZ19,DBLP:conf/nips/TjandraatmadjaA20}, \textit{linear inequalities propagation} \cite{DBLP:conf/nips/ZhangWCHD18,DBLP:conf/aaai/BoopathyWC0D19,DBLP:journals/pacmpl/SinghGPV19}, \textit{Zonotopes} \cite{DBLP:conf/sp/GehrMDTCV18,DBLP:conf/icml/MirmanGV18,DBLP:conf/nips/SinghGMPV18}, \textit{interval bound propagation} (IBP) \cite{DBLP:conf/iccv/GowalDSBQUAMK19}, \textit{dual optimization} \cite{DBLP:conf/icml/WongK18,DBLP:conf/nips/WongSMK18,DBLP:conf/uai/DvijothamSGMK18}, and \textit{semi-definite programming} (SDP) \cite{DBLP:conf/nips/RaghunathanSL18,DBLP:conf/iclr/RaghunathanSL18,DBLP:conf/uai/DvijothamSGGQDK19}. 
Due to the relaxation, the certification results of incomplete methods could be inaccurate, i.e., they may prove that a model is non-robust around a clean sample even if it is indeed robust.
Nevertheless, incomplete methods are better choices considering the expensive computational cost of complete methods on large models.

Furthermore, robustness certification methods usually are extended as certified training methods \cite{DBLP:conf/icml/MirmanGV18,DBLP:conf/iccv/GowalDSBQUAMK19,DBLP:conf/iclr/ZhangCXGSLBH20} to directly improve the robustness of the target model.
However, due to the trade-off between the robustness of a model and its accuracy \cite{DBLP:conf/iclr/MadryMSTV18, DBLP:conf/icml/ZhangYJXGJ19}, certified training methods can cause a decrease in clean accuracy.

\textit{Certified Defenses for RNN models.}
Due to their unique structures and operations, most of the existing certified defenses cannot be applied to RNN models.
To the best of our knowledge, there are only two certified defenses \cite{DBLP:conf/icml/KoLWDWL19,DBLP:conf/ccs/DuJSZLSFYB021} that focus on RNNs.
POPQORN \cite{DBLP:conf/icml/KoLWDWL19} utilizes IBP to certify the RNN model's robustness for the first time. 
Later, Cert-RNN \cite{DBLP:conf/ccs/DuJSZLSFYB021} was proposed, which is based on the Zonotope domain and outperforms the former POPQORN in precision and efficiency.
However, we discover a vulnerability in above works, i.e., they are challenged by multi-frame attacks because they focus on the weaker one-frame attacks, which motivates this work.

\section{Conclusion}
In this paper, we present RNN-Guard, the first certified defense against multi-frame attacks for RNN models.
RNN-Guard adopts the perturb-all-frame strategy for larger perturbation space that captures all potential adversarial examples in multi-frame attacks.
We also introduce a new abstract domain called InterZono and design tighter relaxations to address the precision issue caused by the perturb-all-frame strategy.
We comprehensively evaluate the performance of RNN-Guard across various models and datasets.
The results show that InterZono is more precise than Zonotope while carrying the same time complexity.
Moreover, we extend RNN-Guard as the first certified training method against multi-frame attacks.
The experimental results show that, compared to existing defenses, RNN-Guard is more effective against multi-frame attacks.

\bibliographystyle{IEEEtranS}
\bibliography{main}

\newpage

\begin{IEEEbiographynophoto}{Yunruo Zhang}
received the bachelor’s degree in mathematics from Shandong University, Jinan, China, in 2020. He is currently working toward the Ph.D. degree in the School of Cyber Science and Technology at Shandong University, Qingdao, Shandong, China. His research interests include certified robustness and AI security.
\end{IEEEbiographynophoto}

\begin{IEEEbiographynophoto}{Tianyu Du}
is currently a postdoctoral scholar with the College of Information Science and Technology, Penn State University, advised by Prof. Ting Wang. Prior to that, she obtained her Ph.D. degree in the College of Computer Science and Technology at Zhejiang University, China, on June 2022. Her main research interests include AI security and adversarial machine learning.
\end{IEEEbiographynophoto}

\begin{IEEEbiographynophoto}{Shouling Ji}
(Member, IEEE) received the Ph.D. degree in electrical and computer engineering from the Georgia Institute of Technology, Atlanta, Georgia, and the Ph.D. degree in computer science from Georgia State University, Atlanta, Georgia. He is currently a ZJU 100-Young professor with the College of Computer Science and Technology, Zhejiang University, and a research faculty with the School of Electrical and Computer Engineering, Georgia Institute of Technology. His research interests include AI security, data-driven security, privacy and data analytics. He is a member of the ACM and was the membership chair of the IEEE Student Branch, Georgia State (2012–2013).
\end{IEEEbiographynophoto}

\begin{IEEEbiographynophoto}{Peng Tang}
 received the Ph.D. degree in 2019 from the Beijing University of Posts and Telecommunications, China. He is currently a Associate Professor at Shandong University. His research interests include data privacy and databases.
\end{IEEEbiographynophoto}

\begin{IEEEbiographynophoto}{Shanqing Guo}
received the M.S. degree in computer science from Ocean University, Qingdao, China, in 2003, and the Ph.D. degree in computer science from Nanjing University, Nanjing, China, in 2006. He is currently a professor with the School of Cyber Science and Technology, Shandong University. His research interests include AI security, software security, and system security. He also serves as a program committee member or a reviewer for various international conferences and journals, e.g., ISSRE, ICSME and Computer \& Security.
\end{IEEEbiographynophoto}

\end{document}